\documentclass{bmvc2k}

\usepackage{multirow}
\usepackage{booktabs}

\usepackage{enumitem}

\usepackage{xcolor}

\usepackage{amsfonts}

\newcommand*{\img}[1]{%
    \raisebox{-.07\baselineskip}{%
        \includegraphics[
        height=0.9\baselineskip,
        width=0.9\baselineskip,
        keepaspectratio,
        ]{#1}%
    }%
}

\title{Extract More from Less: \\ 
Efficient 
Fine-Grained
Visual Recognition \\
in Low-Data Regimes
}

\addauthor{Dmitry Demidov}{dmitry.demidov@mbzuai.ac.ae}{1}
\addauthor{Abduragim Shtanchaev}{abduragim.shtanchaev@mbzuai.ac.ae}{1}
\addauthor{Mihail Mihaylov}{mihail.mihaylov@mbzuai.ac.ae}{1}
\addauthor{Mohammad Almansoori}{mohammad.almansoori@mbzuai.ac.ae}{1}

\addinstitution{
 Mohamed bin Zayed University \\ of Artificial Intelligence \\
 UAE, Abu Dhabi \\
}

\runninghead{Dmitry Demidov \lowercase{et al.}} 
{Extract More from Less}

\begin{document}

\maketitle

\begin{abstract}

The emerging task of fine-grained image classification in low-data regimes assumes the presence of low inter-class variance and large intra-class variation along with a highly limited amount of training samples per class.
However, traditional ways of separately dealing with fine-grained categorisation and extremely scarce data may be inefficient under both these harsh conditions presented together.
In this paper,
we present a novel framework, called \mbox{AD-Net}, aiming to enhance deep neural network performance on this challenge by leveraging the power of Augmentation and Distillation techniques. 
Specifically, our approach is designed to refine learned features through self-distillation on augmented samples, mitigating harmful overfitting. We conduct comprehensive experiments on popular fine-grained image classification benchmarks where our \mbox{AD-Net} demonstrates consistent improvement over traditional fine-tuning and 
state-of-the-art low-data techniques. Remarkably, with the smallest data available, our framework shows an outstanding relative accuracy increase of up to 45 \% compared to standard ResNet-50 and up to 27 \% compared to the closest SOTA runner-up.
We emphasise that our approach is practically architecture-independent and adds zero extra cost at inference time.
Additionally, we provide an extensive study on the impact of 
every framework's component, highlighting the importance of each in achieving optimal performance.
Source code and trained models are publicly available at \href{https://github.com/demidovd98/fgic_lowd}{github.com/demidovd98/fgic\_lowd}.

\end{abstract}

\section{Introduction}
\label{sec:intro}

\textbf{Overview.}
Deep learning models, inherently data-hungry, require vast quantities of annotated data for effective training \cite{dosovitskiy2020image}. However acquiring such extensive datasets, particularly with meticulous annotations, is labour-intensive and challenging. 
Following this constraint, the task of low-data image classification assumes that only a highly limited amount of annotated samples per class is available with no other information.
\cite{schmarje2021survey, yang2022survey, xu2021end}.
Despite the advantages of the existing methodologies, they often come with computationally expensive and architecture-dependent extra modules \cite{shu2022improving} and in some cases still demand substantial minimally required data quantities \cite{Wang2021FeatureFV} or expect the input data to be in a specific domain only \cite{10.1007/978-3-031-45676-3_36}.
\begin{figure}[!t]
    \centering
    \includegraphics[width=1.0\linewidth]{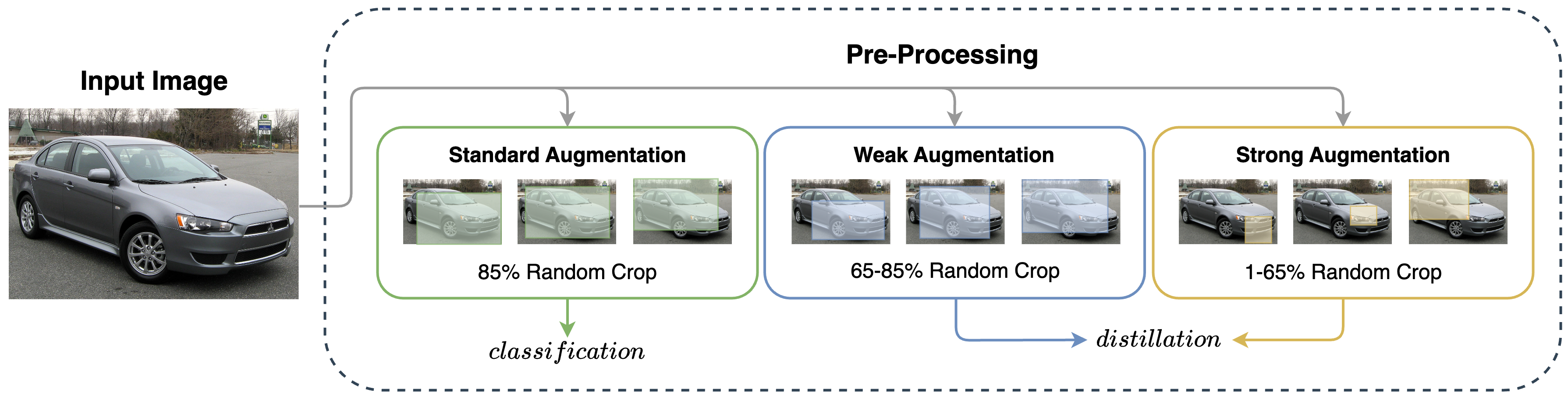}
    %
    \caption{An overview of our proposed pre-processing pipeline. Three random crops of different sizes are generated from each input image and all of them are further augmented with the same set of transformations.
    The largest cropped region is further used for classification and the other two (mid- and small-size) are used as target and source in a distillation objective.
    Only random crop is shown as an applied augmentation for simplicity.}
    \vskip -0.25in
    \label{fig:architecture_augs}
\end{figure}
A potentially helpful traditional way to address this bottleneck is to artificially increase data amounts by applying various data augmentations \cite{simonyan2014very, cubuk2018autoaugment,yun2019cutmix}.
However, our investigation demonstrates that, when applied within the standard transfer learning pipeline \cite{he2016deep}, these approaches are not practically 
efficient under the extremely low amount of data. Another challenging domain within computer vision is fine-grained visual recognition, where the goal is to learn distinctive features for visually or semantically similar sub-classes of the same meta-category \cite{schmarje2021survey, sm-vit}.
This task may 
be difficult even for human annotators 
due to specific domain knowledge requirements \cite{yang2022survey, xu2021end}, making large-scale data acquisition and labelling impractical.
Therefore, the problem of low-data learning is especially important for fine-grained classification tasks where the available data per class is often scarce \cite{yang2022survey, xu2021end}.

\noindent\textbf{Problem Statement.}
Our work focuses on fine-grained image classification (FGIC) under low-data regimes, where both intra-class variation is high and the data is scarce, so conventional solutions may be limited \cite{simonyan2014very}.
This challenge is complicated by the intricate nature of the FGIC task and limited research in transfer learning for low-data problems. Most of the few existing solutions \cite{shu2022improving, LU202225} aim to directly introduce learning constraints without addressing the data variation insufficiency and often at the cost of noticeable time and resource increase.
Therefore, motivated by the under-explored low-data learning domain,
we provide a solution that effectively solves
fine-grained classification problems by utilising a soft and indirectly regularised transfer learning technique in order to mitigate overfitting challenges.

\noindent\textbf{Approach.}
We propose a novel approach which performs two crucial tasks: leverages image augmentations to enrich the feature space and utilises self-distillation to facilitate knowledge refinement among feature layers.
Unlike traditional methods, our single-model approach integrates the classification and teacher-student distillation pipelines in one framework.
Specifically, in our solution, we leverage a self-distillation
objective to maintain close feature distributions between teacher and student outputs for different views of each image, which helps to enhance learned representations. 
Our detailed study offers the following \textit{contributions}:
\begin{itemize}[noitemsep,topsep=0pt, wide=5pt, leftmargin=*]
    \item We propose 
    \underline{an end-to-end low-data framework} which includes two feature distillation branches for model purification
    by soft and implicit regularisation. 
    \item Our
    architecture-independent solution allows a model to 
    \underline{learn and refine representations} by enriching the variability of 
    data through extra augmented image views.
    \item We demonstrate that our approach 
    \underline{achieves state-of-the-art results} on popular FGIC ben-chmarks under extremely low data regimes and helps to reduce overfitting. 
    \item With our framework, the performance gain comes at
    \underline{zero extra cost at inference time}. 
\end{itemize}

In Section \ref{sec:ablation}, we present an extensive study of the architecture design, objective functions, and augmentation types, 
contributing to the evolving landscape of both fine-grained image classification and low-data regimes learning.

\section{Related Work}
\label{sec:related}

\textbf{Fine-grained Image Classification.} 
There exist multiple approaches 
for the fine-grained image classification setting. 
Traditionally,  
pre-trained R-CNNs \cite{zhang2014part} and part detectors \cite{branson2014bird} had been utilised to detect similarities between specific parts of the object in the image. While modern 
approaches tend to utilise end-to-end
architectures
\cite{zhuang2020learning} where a mutual feature vector is obtained from multiple backbones and further used to compare to unique image representations such that the model can distinguish between difficult classes \cite{7410527, 7780410, 10.5555/3454287.3454672}.
Another way to solve the FGIC problem is to focus on robust loss functions \cite{chang2020devil}, for example having both  
a discriminability component which forces all feature channels belonging to the same class to be discriminative, using a
channel-wise attention mechanism, and a diversity component which promotes mutually exclusive channels.
Similar methods also focus on better feature extraction, like aggregating the important tokens from each transformer layer to compensate for local, low-level and middle-level information 
\cite{Wang2021FeatureFV}.
Some of the recent approaches \cite{lagunas2023transfer, chou2022novel} consider utilising the heavy 
Vision Transformer (ViT) \cite{dosovitskiy2020image} architecture and semi-supervised settings, 
however such backbones require a computationally expensive pre-training phase and are noticeably slow at inference time.

\noindent\textbf{Low-Data Learning.} 
The specific task of fine-grained image classification in low-data regimes \cite{horn2017devil} is not widely explored, and, to our knowledge, there exist only a few approaches that attempt to solve it. One of them is using a self-boosting attention mechanism in CNNs to focus on more descriptive parts of the classes \cite{shu2022improving}. However, this FGIC method is designed in an architecture-specific way and can not be extended to other backbones.
There exist other approaches that attempt to improve the training data quality by utilising saliency maps in CNNs to specifically target low-data regime settings. This can be seen in \cite{flores2019saliency} where the authors incorporate a specific attention mechanism in the original image classification pipeline such that the model focuses on the important distinctive parts of the image when classifying similar classes. In \cite{tang2022learning} the authors use a multi-scale feature pyramid and a multi-level attention pyramid on a backbone network to progressively aggregate features from different granular spaces. They further present an attention-guided refinement strategy in collaboration with a multi-level attention pyramid to reduce the uncertainty brought by backgrounds conditioned by limited samples.
Nevertheless, the existing methods are often sub-optimal or architecture-specific and resource-demanding, leaving room for a more elegant solution.

\noindent\textbf{Augmentations.}
\label{sec:efficient_learning}
Recent developments in augmentation strategies for self-supervised models include methods like MultiCrop \cite{caron2020unsupervised} which employ a mix of views with different resolutions, enhancing encoding space variability without increasing memory or compute requirements.
Another powerful augmentation is ScaleMix \cite{wang2022importance} which generates new views of an image by mixing two views of potentially different scales using binary masking. Similar to ScaleMix, \mbox{CutMix \cite{yun2019cutmix}} operates exclusively on views from the same image, producing a single view of standard size, however, the latter also approximates benefits of multiple crops without requiring separate processing of small crops. 
Both MultiCrop and ScaleMix introduce extra variance to the encoding space, enriching the representations. 

\noindent\textbf{Distillation.}
Another recent approach utilised for efficient learning is model distillation during training. Self-supervised learning methodologies have increasingly embraced distillation objectives as a central principle for robust representation learning in the absence of explicit labels \cite{jaiswal2021survey, mazumder2021fair}. Empirical studies in the literature, including works such as \cite{caron2021emerging, grill2020bootstrap, zbontar2021barlow}, demonstrate reasonable performance, particularly in the challenging paradigm of zero-shot learning \cite{sultana2022selfdistilled}. Drawing motivation from this paradigm, we attempt to integrate the distillation techniques into our approach to target the challenging scenarios
of limited data availability.

\section{Method}
\label{sec:method}

\subsection{Baseline Framework}

\textbf{Transfer Learning.}
In traditional transfer learning, a backbone is usually pre-trained on a huge and visually diverse dataset (such as ImageNet \cite{imagenet_v1}) and is further fine-tuned on a smaller fine-grained downstream 
task in order to 
transfer generic low- and mid-level features learned on the large and diverse dataset.
This type of initialisation is especially beneficial in the case of using 
smaller fine-grained datasets \cite{lagunas2023transfer, darvish2022towards}.
Adopting standard terminology in transfer learning, we define the source domain \(D_S\) with data \(X_S\) 
and the target domain \(D_T\) with data \(X_T\).
Here, \(X = \{x_1, ..., x_n\}\) represents the feature set within domain-specific feature spaces \(\chi_S\) and \(\chi_T\), respectively.
The objective in this context is to predict the label \(y'\) for a new instance \(x'\) in the target domain, utilising a label space \(Y\) and learning a prediction function \(f: \chi_T \rightarrow Y\). This function is derived from training pairs \((x_i, y_i)\), where \(x_i \in X_S \cup X_T\) and \(y_i \in Y\), effectively mapping the learned features to the desired outputs.

\noindent\textbf{Limitations.}
Although the traditional fine-tuning strategy is useful for standard datasets with relative abundance of labelled samples, it may be sub-optimal on tasks with highly limited data.
In low-data regimes, this practically leads to dramatic overfitting to the few given samples per class.
Moreover, previous works on low-data learning \cite{shu2022improving,Wang2021FeatureFV} have demonstrated that the unconstrained baselines mostly focus on the most obvious but less distinctive visual features. Taking this into account, we address these challenges of the low-data setting with our proposed approach, which is able to reduce harmful overfitting by refining learned representations through self-distillation 
on multiple augmented
views of the input image.

\subsection{Our Approach}
\label{sec:approach}

\begin{figure*}[!t]
    \centering
    \includegraphics[width=1.0\linewidth]{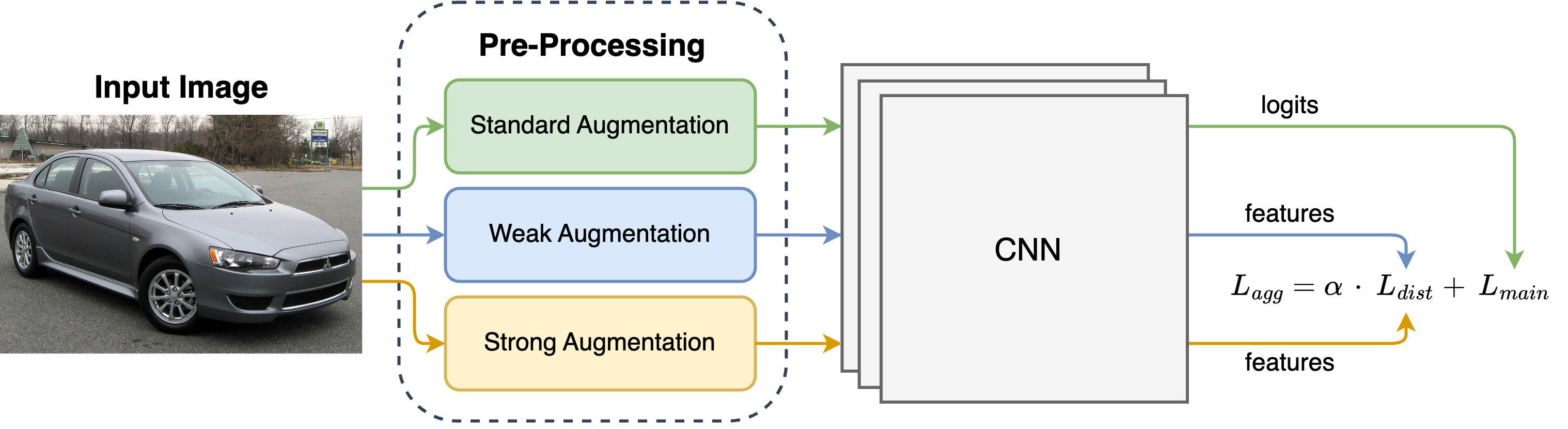}
    \caption{A proposed architecture. A random crop is applied as an input augmentation for images which are subsequently fed into the same model with shared weights. Further, we compare the feature outputs of different crops with a distillation loss $\mathcal{L}_{dist}$. The aggregated loss $\mathcal{L}_{agg}$ is a combination of the traditional cross-entropy loss $\mathcal{L}_{main}$ and $\mathcal{L}_{dist}$.
    }
    \vskip -0.15in
    \label{fig:architecture}
\end{figure*}

\textbf{Overall Architecture.}
In this paper, we propose a framework, called AD-Net, which aims to achieve the enhancement of deep neural network performance through the combination of Augmentation and Distillation techniques. 
Such a combination leads to substantial quality improvement of learned information per image, which allows for the neural network to learn better representations with fewer images.
Our primary focus is on the application of self-distillation techniques, incorporating an additional distillation loss applied to augmented input samples to facilitate knowledge enhancement within the network.
More specifically, we explore the utilisation of various data augmentations to further improve the model's robustness and experiment with different distillation loss functions and components, providing a comprehensive analysis of their effects on the model's performance.
Taking this into account, our proposed framework incorporates a multi-branch configuration, functioning in a manner akin to Siamese networks \cite{jaiswal2021survey, mazumder2021fair}. This design choice allows for weight sharing between branches, enhancing computational efficiency and facilitating the consolidation of learned features. Despite the shared weights, each branch is uniquely
controlled by distinct loss objectives, emphasising their individual contributions to the overall learning process. Specifically, the branches can be delineated as the classification branch and the distillation branches for simplicity. Figure \ref{fig:architecture} provides an overview of our proposed framework.

\noindent\textbf{Classification Branch.}
\label{sec:global_branch}
This branch replicates the standard supervised training process, which follows the conventional fine-tuning procedure \cite{9298575} with an ordinary 
classification objective. Full-sized images with traditional data augmentation techniques applied are fed into this branch in order to separately obtain the class prediction distribution.
\\

\noindent\textbf{Distillation Branches.}
\label{augmented branch}
This part of the architecture performs knowledge distillation with two differently augmented views of the input image. 
More specifically, one ``source" branch and one ``target" branch are utilised to generate corresponding source and target representations of the original images, which are further used for self-distillation \cite{deit, chen2020big-Self-Supervised-Models}. This encourages the model to do feature refinement towards more useful class-specific patterns. 
In more detail, both branches utilise the same shared-weight 
backbone from the main classification branch, but it is now fed with random mid-sized and small crops of the original image.
Specifically, following \cite{wang2022importance}, we randomly sample two categories of patches, larger target crops ranging between 65-85\% of the original image content and smaller source crops with an area range of 1-65\% in order to capture 
fine-grained class-specific patterns \cite{caron2021emerging}. 
This kind of image space limitation can help to increase data variation and introduce implicit feature pruning via information dropout \cite{DeiT-III-Revenge-of-the-ViT}.
Next, all sampled regions are resized to the fixed resolution and augmented with the same set of data augmentations used for the classification branch.
Thus, for each input image, one cropped region from both categories is fed into the corresponding branches of the model to obtain two different representations.
Finally, inspired by \cite{deit}, a self-distillation technique is applied to the corresponding feature outputs of previously sampled image crops, which encourages the model to learn better class-specific patterns. This helps to overcome
quick
overfitting towards explicit but weak features.

\noindent\textbf{Loss Formulation.}
In our approach, we deploy a combination of two distinct loss functions to optimise learning efficiency. The primary classification objective, denoted as $\mathcal{L}_{main}$, explicitly facilitates the probability distribution prediction towards a pre-defined class. Concurrently, secondary distillation objective $\mathcal{L}_{dist}$ serves as an implicit regularisation mechanism \cite{wang2022importance}.
More specifically, the classification loss is applied to the output logits of the classification branch, while the distillation loss is applied to the feature maps derived from the target and source distillation branches (refer to Section \ref{augmented branch}).
For our main classification objective, we employ
traditionally used Cross-Entropy loss
\cite{mao2023crossentropy, hui2021evaluation, zhang2018generalized}:
\vskip -0.1in
\begin{equation}
    \mathcal{L}_{main} = -\frac{1}{N} \sum_{n=1}^{N} \sum_{i}^{C} y_{ni} \log(\hat{y}_{ni})
\end{equation}
\vskip -0.05in
where $y_{ni}$ is the ground truth label and  $\hat{y}_{ni}$ is the prediction of the model.

Meanwhile, for the distillation objective, we utilise 
Kullback–Leibler divergence
\cite{kim2021comparing}:
\vskip -0.1in
\begin{equation}
\mathcal{L}_{dist} = \frac{1}{N} \sum_{n=1}^{N} \sum_{i}^{C} P^{t}_{ni} \log\left(\frac{P^{t}_{ni}}{Q^{s}_{ni}}\right)
\end{equation}
\vskip -0.05in
where $P^{t}_{ni}$ and $Q^{s}_{ni}$ are feature distributions from
target and source distillation branches.

\noindent In our experiments we apply the softmax function on feature outputs to convert them into normalised probability distributions.
The main intuition behind this decision is to enforce the model to produce similar feature distributions 
for different regions of the same image.
Finally, following a widely accepted way of 
objectives combination \cite{shu2022improving, 10.5555/3618408.3619150, 10.1007/978-3-031-45676-3_36}, our aggregated loss function ${L}_{agg}$, consisting of the classification and distillation components, is:
\vskip -0.15in
\begin{equation}
\mathcal{L}_{agg} =  \mathcal{L}_{main} + \alpha \cdot \mathcal{L}_{dist}
\label{eq:final_loss}
\end{equation}
\vskip -0.03in
where $\alpha$ is a hyper-parameter determining the weight of the distillation loss in the final objective (for more details and ablation analysis refer to Section \ref{sec:ablation}).

\section{Experiments}
\label{sec:experiments}

\subsection{Experimental Details}

\textbf{Datasets.}
\label{sec:datasets}
We explore the properties of our approach on the following popular FGIC datasets: CUB-200-2011 \cite{WahCUB_200_2011}, Stanford Cars \cite{Krause2013CollectingAL}, FGVC-Aircraft \cite{maji13fine-grained}.
The following datasets have been chosen due to their balanced class sizes and a similar number of images per class, which allow for a more fair evaluation (for details refer to Appendix \ref{app:datasets}). 
In order to imitate low-data regimes, we strictly follow the sampling proposed in \cite{shu2022improving}, where
samples are randomly drawn from the training set providing 10\%, 15\%, 30\%, and 50\% percentages of 
data.

\noindent\textbf{Baselines.}
In order to provide a fair comparison, in our experiments we utilise methods based on popular CNN and ViT architectures, where for all models (including ours) we perform transfer learning by fine-tuning.
For the 
vanilla baselines
we also consider two setups: traditional fine-tuning and advanced fine-tuning with our proposed training recipe designed for low-data learning.
For the comparison, we first consider methods specifically designed for the fine-grained classification task, such as: Full Bilinear Pooling (FBP) \cite{7410527}, Compact Bilinear Pooling with Tensor Sketch projection (CBP-TS) \cite{7780410}, Hierarchical Bilinear Pooling (HBP) \cite{10.1007/978-3-030-01270-0_35}, Deep Bilinear Transformation (DBTNet-50) \cite{10.5555/3454287.3454672}, and FFVT \cite{Wang2021FeatureFV}.
As a main competitor, we evaluate the first and currently the only specifically designed approach for low-data in FGIC \cite{shu2022improving}, which is a CNN-based model with a self-boosting attention mechanism, by preserving the authors' codebase and training hyper-parameters.
For our approach, we similarly use a family of ResNet models \cite{he2016deep} as a main backbone
along with such other popular baselines as: GoogLeNet \cite{7298594}, DenseNet \cite{Huang2016DenselyCC}, Inception v3 \cite{7780677}, and ViT \cite{dosovitskiy2020image}.
All considered models were standardly pre-trained on ImageNet and further fine-tuned on the above-mentioned fine-grained datasets for both vanilla baselines and our AD-Net.

\noindent\textbf{Implementation Details.}
Compared to the traditional fine-tuning procedure, initially designed for a sufficient amount of samples per class, our proposed training recipe includes several adaptations for the low-data regimes (for application guidelines
refer to \mbox{Appendix \ref{app:qna})}. 
First, following a common idea of increasing the main learning rate value $lr$ by a fixed ratio for the last classification layer $lr_{cls}$, we automatically update this ratio depending on the percentage of the available data $\mathcal{D}_{aval}$ with the following rule:
\vskip -0.1in
\begin{equation}
    \label{eq:lr_ratio}
   lr_{cls} = lr \cdot \biggl( round \bigl( \frac{100}{\mathcal{D}_{aval}} \cdot 2 \bigr) / 2 \biggr) .
\end{equation}
\vskip -0.02in
This assumption is valid in the simulated setup due to similar dataset sizes, but it can be further extended by replacing the available data percentage with the number of images per class adjusted to the number of classes.
Additionally, we utilise a learning rate scheduler with downscaling $lr$ for 5 iterations $iter \in [1:5]$ by $0.1$ at each $(steps_{max} \cdot 0.5^{iter})$ training step.
For other implementation details and training hyper-parameters refer to Appendix \ref{app:implementation}.

\begin{table*}[t]
\centering
\caption{Comparison of different approaches 
using various percentages of the data on three popular FGIC datasets. Our proposed solution achieves consistent improvement in performance over other methods across low data settings. 
Best results are highlighted in bold.}
\label{tab:results_main}
\begin{tabular}{@{}c|l|cccc@{}}
\toprule
\multirow{2}{*}{Dataset} &
  \multicolumn{1}{c|}{\multirow{2}{*}{Method}} &
  \multicolumn{4}{c}{Training data percentage} \\ \cmidrule(l){3-6}
 &
  \multicolumn{1}{c|}{} &
  \multicolumn{1}{@{\hspace{1.45em}}c@{\hspace{1.45em}}|}{10\%} &
  \multicolumn{1}{@{\hspace{1.45em}}c@{\hspace{1.45em}}|}{15\%} &
  \multicolumn{1}{@{\hspace{1.45em}}c@{\hspace{1.45em}}|}{30\%} &
  \multicolumn{1}{@{\hspace{1.45em}}c@{\hspace{1.45em}}}{50\%}
  \\ \midrule
\multirow{3}{*}{CUB-200-2011 \cite{WahCUB_200_2011}
} &
  ResNet-50 
  &
  \multicolumn{1}{c|}{36.99} &
  \multicolumn{1}{c|}{48.88} &
  \multicolumn{1}{c|}{62.60} &
  \multicolumn{1}{c}{73.23}
  \\
 &
  FBP
  &
  \multicolumn{1}{c|}{37.88} &
  \multicolumn{1}{c|}{49.12} &
  \multicolumn{1}{c|}{63.27} &
  \multicolumn{1}{c}{73.70} 
  \\
 &
  CBP-TS
  &
  \multicolumn{1}{c|}{37.12} &
  \multicolumn{1}{c|}{47.82} &
  \multicolumn{1}{c|}{62.24} &
  \multicolumn{1}{c}{72.37}
  \\
 &
  HBP
  &
  \multicolumn{1}{c|}{38.57} &
  \multicolumn{1}{c|}{50.12} &
  \multicolumn{1}{c|}{63.86} &
  \multicolumn{1}{c}{74.18} 
  \\
 &
  DBTNet-50 
  &
  \multicolumn{1}{c|}{37.67} &
  \multicolumn{1}{c|}{49.52} &
  \multicolumn{1}{c|}{63.16} &
  \multicolumn{1}{c}{73.28} 
  \\  
 &
  SAM (ResNet-50) 
  &
  \multicolumn{1}{c|}{40.24} &
  \multicolumn{1}{c|}{52.05} &
  \multicolumn{1}{c|}{64.07} &
  \multicolumn{1}{c}{73.92}
  \\
 &
  SAM (FBP)
  &
  \multicolumn{1}{c|}{41.83} &
  \multicolumn{1}{c|}{52.35} &
  \multicolumn{1}{c|}{65.19} &
  \multicolumn{1}{c}{74.54}
  \\  
 &
  \textit{Ours (ResNet-50)} &
  \multicolumn{1}{c|}{\textbf{47.51}} &
  \multicolumn{1}{c|}{\textbf{60.08}} &
  \multicolumn{1}{c|}{\textbf{71.11}} &
  \multicolumn{1}{c}{\textbf{77.67}}
  \\ \midrule
\multirow{3}{*}{Stanford Cars \cite{Krause2013CollectingAL}
} &
  ResNet-50
  &
  \multicolumn{1}{c|}{37.45} &
  \multicolumn{1}{c|}{53.01} &
  \multicolumn{1}{c|}{75.26} &
  \multicolumn{1}{c}{83.56}
  \\
 &
  FBP
  &
  \multicolumn{1}{c|}{40.13} &
  \multicolumn{1}{c|}{55.07} &
  \multicolumn{1}{c|}{76.42} &
  \multicolumn{1}{c}{85.10}
  \\
 &
  CBP-TS
  &
  \multicolumn{1}{c|}{37.77} &
  \multicolumn{1}{c|}{54.87} &
  \multicolumn{1}{c|}{75.51} &
  \multicolumn{1}{c}{84.80}
  \\
 &
  HBP
  &
  \multicolumn{1}{c|}{40.02} &
  \multicolumn{1}{c|}{55.82} &
  \multicolumn{1}{c|}{76.81} &
  \multicolumn{1}{c}{85.31}
  \\
 &
  DBTNet-50 
  &
  \multicolumn{1}{c|}{39.48} &
  \multicolumn{1}{c|}{55.24} &
  \multicolumn{1}{c|}{76.52} &
  \multicolumn{1}{c}{86.52}
  \\  
 &
  SAM (ResNet-50)
  &
  \multicolumn{1}{c|}{39.96} &
  \multicolumn{1}{c|}{55.02} &
  \multicolumn{1}{c|}{76.69} &
  \multicolumn{1}{c}{84.85}
  \\
 &
  SAM (FBP) 
  &
  \multicolumn{1}{c|}{43.19} &
  \multicolumn{1}{c|}{57.42} &
  \multicolumn{1}{c|}{77.63} &
  \multicolumn{1}{c}{85.71}
  \\  
 &
  \textit{Ours (ResNet-50)} &
  \multicolumn{1}{c|}{\textbf{55.09}} &
  \multicolumn{1}{c|}{\textbf{67.42}} &
  \multicolumn{1}{c|}{\textbf{81.53}} &
  \multicolumn{1}{c}{\textbf{87.41}}
  \\ \midrule
\multirow{3}{*}{FGVC-Aircraft \cite{maji13fine-grained}
} &
  ResNet-50
  &
  \multicolumn{1}{c|}{43.52} &
  \multicolumn{1}{c|}{53.17} &
  \multicolumn{1}{c|}{71.32} &
  \multicolumn{1}{c}{78.61}
  \\
 &
  FBP
  &
  \multicolumn{1}{c|}{45.16} &
  \multicolumn{1}{c|}{55.06} &
  \multicolumn{1}{c|}{72.12} &
  \multicolumn{1}{c}{79.93} 
  \\
 &
  CBP-TS
  &
  \multicolumn{1}{c|}{44.63} &
  \multicolumn{1}{c|}{54.79} &
  \multicolumn{1}{c|}{71.32} &
  \multicolumn{1}{c}{79.60}
  \\
 &
  HBP
  &
  \multicolumn{1}{c|}{45.28} &
  \multicolumn{1}{c|}{56.12} &
  \multicolumn{1}{c|}{72.58} &
  \multicolumn{1}{c}{81.47}
  \\
 &
  DBTNet-50 
  &
  \multicolumn{1}{c|}{45.35} &
  \multicolumn{1}{c|}{56.36} &
  \multicolumn{1}{c|}{73.06} &
  \multicolumn{1}{c}{81.26}
  \\  
 &
  SAM (ResNet-50)
  &
  \multicolumn{1}{c|}{46.73} &
  \multicolumn{1}{c|}{56.02} &
  \multicolumn{1}{c|}{72.59} &
  \multicolumn{1}{c}{79.21}
  \\
 &
  SAM (FBP)
  &
  \multicolumn{1}{c|}{47.97} &
  \multicolumn{1}{c|}{57.47} &
  \multicolumn{1}{c|}{73.43} &
  \multicolumn{1}{c}{80.86}
  \\  
 &
  \textit{Ours (ResNet-50)} &
  \multicolumn{1}{c|}{\textbf{55.81}} &
  \multicolumn{1}{c|}{\textbf{62.59}} &
  \multicolumn{1}{c|}{\textbf{74.44}} &
  \multicolumn{1}{c}{\textbf{81.73}}
  \\ \bottomrule
\end{tabular}
\vskip -0.1in
\end{table*}

\subsection{Results and Analysis}

\textbf{Quantitative Analysis.}
In order to thoroughly investigate our proposed method, we conduct experiments on different label proportions of the popular FGIC benchmarks and further compare the results with the state-of-the-art approaches.
Results in Table \ref{tab:results_main} demonstrate that our self-distillation component significantly improves the performance of the vanilla ResNet-50 across all of the sampled percentages on all considered datasets,
while also outperforming other SOTA methods in the low-data settings.
Specifically, with the smallest data available, our framework shows an outstanding relative accuracy increase of up to 45 \% compared to standard ResNet-50 and up to 27 \%, compared to the closest SOTA runner-up. 
Our solution also outperforms the specifically designed for low-data fine-grained classification SAM framework \cite{shu2022improving},
while not introducing any extra resource-consuming and architecture-specific module.
It can be observed that the most significant improvement is achieved with the smallest number of samples per class available, while the relative performance increase goes down with more data used for training.
This could be explained by the overall data saturation on larger data percentages available, which is the same behaviour for every considered method.
Nevertheless, the benchmark analysis proves the ability of our solution to adapt to the setups with limited data by performing feature refinement through self-distillation on the existing samples. For a detailed comparison including other low-performing baselines refer to Appendix \ref{app:results_full}.
Additionally, we demonstrate the high transferability of our approach by utilising it on top of most of the popular CNN- and ViT-based backbones (see Appendix \ref{app:transferability} for details). The absolute improvement varies between 3-10 \% showcasing that distilling local augmentations
is practically an architecture-independent technique.

\begin{figure}[!h]
    \centering
    \includegraphics[width=1.0\linewidth]{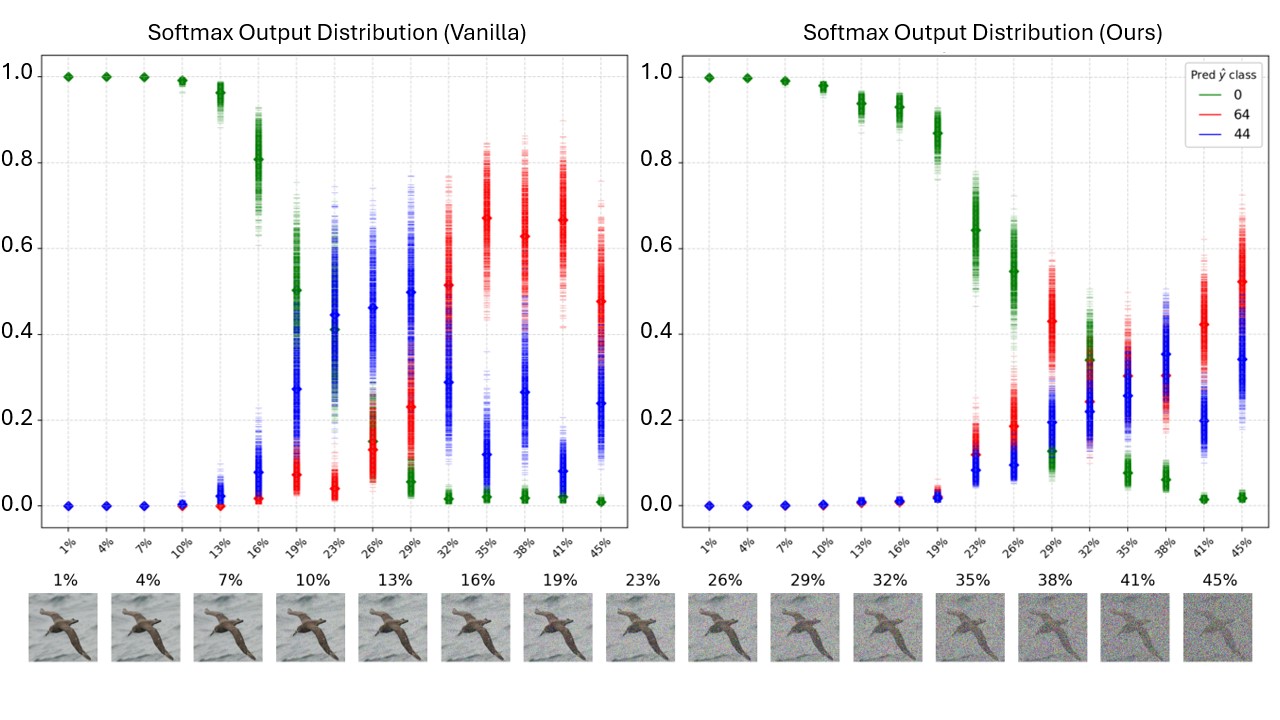}
    \vskip -0.3in
    \caption{Model prediction probability distribution over 1000 forward passes with Monte Carlo Dropout. X-axis stands for the amount of Gaussian noise added (see bottom row) and Y-axis is the probability of a predicted class.
    Each dash (\img{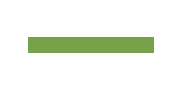}, \img{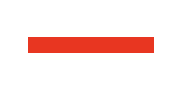}, \img{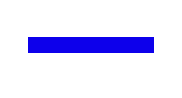}) 
    represents a single model's prediction, where green is the correct class and others are the Top-2 following classes.
    Each diamond (\img{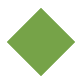}, \img{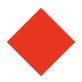}, \img{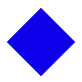}) 
    stands for the mean probability over all predictions. 
    }
    \label{fig:UQ_1}
\end{figure}

\noindent\textbf{Qualitative Analysis.}
In order to analyse the motivation behind the significant performance improvement with our approach, we 
examine the model's performance under different levels of noise added to the input images. 
Specifically, we evaluate the model's prediction uncertainty by using Monte Carlo Dropout \cite{MCdropout}. 
We first sample a random image from the validation set and create a set of 1000 perturbed copies of the image by adding Gaussian noise with varying intensity. 
We then perform a forward pass using vanilla ResNet and our AD-Net for each image in the set (comparison illustrated in \mbox{Figure \ref{fig:UQ_1})}.
We can observe that while the vanilla's predictions have a high false positive rate on noisy samples, the predictions of our AD-Net have lower dispersion and are more stable due to more robust purified features.
For more qualitative analysis and visualisations refer to Appendix \ref{app:anal_qual} and \ref{app:visuals}.

\subsection{Ablation Study}
\label{sec:ablation}

\noindent\textbf{Architecture Design.}
In order to justify the chosen multi-branch pipeline, we provide an architecture design ablation for our framework (see Table \ref{tab:ablation_architecture}). 
Specifically, we first measure the performance of the standard classification branch with the naive fine-tuning procedure. 
Next, we check the performance of this model with our advanced low-data training recipe, which includes an adaptive learning rate ratio for the final classification layer and a scheduler.
Further, we add a single distillation branch which is fed with a small cropped region and then feature distillation is performed between the classification and distillation feature outputs.
Finally, we show the performance of our best solution with one target and one source distillation branches which are fed with corresponding larger target crop and smaller source crop in order to perform a separate self-distillation procedure.
\begin{table}[!ht]
\centering
\caption{Ablation results on our framework architecture design and training recipe.
The experiments are performed on ResNet-50 and the CUB dataset with 10\% of the 
data.
}
\label{tab:ablation_architecture}
\begin{tabular}{@{}c|c|l|c@{\hspace{0.5em}}}
\toprule
Distillation                   & \multicolumn{1}{l|}{Our recipe} & \multicolumn{1}{c|}{Augmentation} & Acc, \%            \\ \midrule
\multirow{2}{*}{-}             & $\times$                               & Basic                           & 36.99          \\
                               & \checkmark	                               & Basic                           & 40.05          \\ \midrule
\multirow{3}{*}{Single-branch} & \checkmark	                               & ScaleMix                        
& 45.98          \\
                               & \checkmark	                               & MultiCrop                       
                               & 46.22          \\
                               & \checkmark	                               & Basic                              & 47.09          \\ \midrule
\multirow{4}{*}{Double-branch} & \checkmark	                               & ScaleMix                           
& 46.11          \\
                               & \checkmark	                               & MultiCrop                         
                               & 45.91          \\
                               & \checkmark	                               & AsymAug                           
                               & 47.18          \\
(Ours)                       & \checkmark	                               & Basic                              & \textbf{47.51} \\ \bottomrule
\end{tabular}
\end{table}

Interestingly, application of advanced augmentation techniques, such as ScaleMix \cite{wang2022importance}, \mbox{MultiCrop \cite{caron2020unsupervised}}, and AsymAug \cite{wang2022importance}, within our self-distillation branches demonstrated some gains, but is not superior. 
We assume that this could be explained by the strong nature of these augmentations, which may not be as efficient due to the harsh perturbations.
For more ablation analysis and limitations refer to Appendix \ref{app:anal_ablation}, \ref{app:limitations}, and \ref{app:extra_experiments}.

\section{Conclusion}
\label{sec:conclusion}

Our proposed AD-Net framework demonstrates exceptional results on the fine-grained image classification task under low-data regimes, which typically requires the identification of local and subtle class-specific features.
This is achieved by integrating advanced augmentation techniques and a distinctive self-distillation strategy, wherein a single model processes both the original image and its transformed views. This multi-input approach enhances the model's ability to recognise and differentiate intricate features unique to each class. 

Through comprehensive experiments, we have established that our model surpasses existing state-of-the-art techniques by a significant margin, particularly excelling in scenarios with the most limited data (see Appendix \ref{app:visuals}). 
Specifically, with the smallest data available, our AD-Net shows an outstanding relative accuracy increase of up to 45 \% compared to standard ResNet-50 and up to 27 \% compared to the closest SOTA runner-up.
This success is attributed to the synergistic combination of augmentations and a tailored objective function, which collectively improve the model's learning quality.
In detail, we explain this noticeable performance gain by the effect of the self-distillation objective, which provides a separate and more detailed source of information by enforcing feature space alignment for the augmented views of the same input image.
Such a setup allows our AD-Net to avoid quick overfitting typical for the standard fine-tuning procedure (refer to Appendix \ref{app:anal_qual}).

We also emphasise that our proposed framework is practically architecture-independent since it requires only the final feature representation output 
from the utilised baseline.
Along with zero extra costs at inference time,
the above-mentioned benefits make our AD-Net framework highly suitable for low-data fine-grained image classification problems.

\bibliography{main}

\begin{thebibliography}{58}
\providecommand{\natexlab}[1]{#1}
\providecommand{\url}[1]{\texttt{#1}}
\expandafter\ifx\csname urlstyle\endcsname\relax
  \providecommand{\doi}[1]{doi: #1}\else
  \providecommand{\doi}{doi: \begingroup \urlstyle{rm}\Url}\fi

\bibitem[Branson et~al.(2014)Branson, Van~Horn, Belongie, and Perona]{branson2014bird}
Steve Branson, Grant Van~Horn, Serge Belongie, and Pietro Perona.
\newblock Bird species categorization using pose normalized deep convolutional nets.
\newblock \emph{arXiv preprint arXiv:1406.2952}, 2014.

\bibitem[Caron et~al.(2020{\natexlab{a}})Caron, Misra, Mairal, Goyal, Bojanowski, and Joulin]{caron2020unsupervised}
Mathilde Caron, Ishan Misra, Julien Mairal, Priya Goyal, Piotr Bojanowski, and Armand Joulin.
\newblock Unsupervised learning of visual features by contrasting cluster assignments.
\newblock \emph{Advances in neural information processing systems}, 33:\penalty0 9912--9924, 2020{\natexlab{a}}.

\bibitem[Caron et~al.(2020{\natexlab{b}})Caron, Misra, Mairal, Goyal, Bojanowski, and Joulin]{swav}
Mathilde Caron, Ishan Misra, Julien Mairal, Priya Goyal, Piotr Bojanowski, and Armand Joulin.
\newblock Unsupervised learning of visual features by contrasting cluster assignments.
\newblock In \emph{Proceedings of the 34th International Conference on Neural Information Processing Systems}, NIPS '20, Red Hook, NY, USA, 2020{\natexlab{b}}. Curran Associates Inc.
\newblock ISBN 9781713829546.

\bibitem[Caron et~al.(2021)Caron, Touvron, Misra, Jégou, Mairal, Bojanowski, and Joulin]{caron2021emerging}
Mathilde Caron, Hugo Touvron, Ishan Misra, Hervé Jégou, Julien Mairal, Piotr Bojanowski, and Armand Joulin.
\newblock Emerging properties in self-supervised vision transformers, 2021.

\bibitem[Chang et~al.(2020)Chang, Ding, Xie, Bhunia, Li, Ma, Wu, Guo, and Song]{chang2020devil}
Dongliang Chang, Yifeng Ding, Jiyang Xie, Ayan~Kumar Bhunia, Xiaoxu Li, Zhanyu Ma, Ming Wu, Jun Guo, and Yi-Zhe Song.
\newblock The devil is in the channels: Mutual-channel loss for fine-grained image classification.
\newblock \emph{IEEE Transactions on Image Processing}, 29:\penalty0 4683--4695, 2020.

\bibitem[Chen et~al.(2020)Chen, Kornblith, Swersky, Norouzi, and Hinton]{chen2020big-Self-Supervised-Models}
Ting Chen, Simon Kornblith, Kevin Swersky, Mohammad Norouzi, and Geoffrey Hinton.
\newblock Big self-supervised models are strong semi-supervised learners, 2020.

\bibitem[Chou et~al.(2022)Chou, Lin, and Kao]{chou2022novel}
Po-Yung Chou, Cheng-Hung Lin, and Wen-Chung Kao.
\newblock A novel plug-in module for fine-grained visual classification.
\newblock \emph{arXiv preprint arXiv:2202.03822}, 2022.

\bibitem[Cubuk et~al.(2018)Cubuk, Zoph, Mane, Vasudevan, and Le]{cubuk2018autoaugment}
Ekin~D Cubuk, Barret Zoph, Dandelion Mane, Vijay Vasudevan, and Quoc~V Le.
\newblock Autoaugment: Learning augmentation policies from data.
\newblock \emph{arXiv preprint arXiv:1805.09501}, 2018.

\bibitem[Cubuk et~al.(2020)Cubuk, Zoph, Shlens, and Le]{cubuk2020randaugment}
Ekin~D Cubuk, Barret Zoph, Jonathon Shlens, and Quoc~V Le.
\newblock Randaugment: Practical automated data augmentation with a reduced search space.
\newblock In \emph{Proceedings of the IEEE/CVF conference on computer vision and pattern recognition workshops}, pages 702--703, 2020.

\bibitem[Darvish et~al.(2022)Darvish, Pouramini, and Bahador]{darvish2022towards}
Mahdi Darvish, Mahsa Pouramini, and Hamid Bahador.
\newblock Towards fine-grained image classification with generative adversarial networks and facial landmark detection.
\newblock In \emph{2022 International Conference on Machine Vision and Image Processing (MVIP)}, pages 1--6. IEEE, 2022.

\bibitem[Demidov. et~al.(2023)Demidov., Sharif., Abdurahimov., Cholakkal., and Khan.]{sm-vit}
Dmitry Demidov., Muhammad Sharif., Aliakbar Abdurahimov., Hisham Cholakkal., and Fahad Khan.
\newblock Salient mask-guided vision transformer for fine-grained classification.
\newblock In \emph{Proceedings of the 18th International Joint Conference on Computer Vision, Imaging and Computer Graphics Theory and Applications (VISIGRAPP 2023) - Volume 4: VISAPP}, pages 27--38. INSTICC, SciTePress, 2023.
\newblock ISBN 978-989-758-634-7.
\newblock \doi{10.5220/0011611100003417}.

\bibitem[Demidov et~al.(2024)Demidov, Al~Majzoub, Kumar, and Khan]{10.1007/978-3-031-45676-3_36}
Dmitry Demidov, Roba Al~Majzoub, Amandeep Kumar, and Fahad Khan.
\newblock Distilling local texture features for colorectal tissue classification in low data regimes.
\newblock In Xiaohuan Cao, Xuanang Xu, Islem Rekik, Zhiming Cui, and Xi~Ouyang, editors, \emph{Machine Learning in Medical Imaging}, pages 357--366, Cham, 2024. Springer Nature Switzerland.
\newblock ISBN 978-3-031-45676-3.

\bibitem[Deng et~al.(2009)Deng, Dong, Socher, Li, Li, and Fei-Fei]{imagenet_v1}
Jia Deng, Wei Dong, Richard Socher, Li-Jia Li, Kai Li, and Li~Fei-Fei.
\newblock Imagenet: A large-scale hierarchical image database.
\newblock In \emph{2009 IEEE Conference on Computer Vision and Pattern Recognition}, pages 248--255, 2009.
\newblock \doi{10.1109/CVPR.2009.5206848}.

\bibitem[Dosovitskiy et~al.(2020)Dosovitskiy, Beyer, Kolesnikov, Weissenborn, Zhai, Unterthiner, Dehghani, Minderer, Heigold, Gelly, et~al.]{dosovitskiy2020image}
Alexey Dosovitskiy, Lucas Beyer, Alexander Kolesnikov, Dirk Weissenborn, Xiaohua Zhai, Thomas Unterthiner, Mostafa Dehghani, Matthias Minderer, Georg Heigold, Sylvain Gelly, et~al.
\newblock An image is worth 16x16 words: Transformers for image recognition at scale.
\newblock \emph{arXiv preprint arXiv:2010.11929}, 2020.

\bibitem[Flores et~al.(2019)Flores, Gonzalez-Garcia, van~de Weijer, and Raducanu]{flores2019saliency}
Carola~Figueroa Flores, Abel Gonzalez-Garcia, Joost van~de Weijer, and Bogdan Raducanu.
\newblock Saliency for fine-grained object recognition in domains with scarce training data.
\newblock \emph{Pattern Recognition}, 94:\penalty0 62--73, 2019.

\bibitem[Gal and Ghahramani(2016)]{MCdropout}
Yarin Gal and Zoubin Ghahramani.
\newblock Dropout as a bayesian approximation: Representing model uncertainty in deep learning, 2016.

\bibitem[Gao et~al.(2016)Gao, Beijbom, Zhang, and Darrell]{7780410}
Yang Gao, Oscar Beijbom, Ning Zhang, and Trevor Darrell.
\newblock Compact bilinear pooling.
\newblock In \emph{2016 IEEE Conference on Computer Vision and Pattern Recognition (CVPR)}, pages 317--326, 2016.
\newblock \doi{10.1109/CVPR.2016.41}.

\bibitem[Grill et~al.(2020)Grill, Strub, Altché, Tallec, Richemond, Buchatskaya, Doersch, Pires, Guo, Azar, Piot, Kavukcuoglu, Munos, and Valko]{grill2020bootstrap}
Jean-Bastien Grill, Florian Strub, Florent Altché, Corentin Tallec, Pierre~H. Richemond, Elena Buchatskaya, Carl Doersch, Bernardo~Avila Pires, Zhaohan~Daniel Guo, Mohammad~Gheshlaghi Azar, Bilal Piot, Koray Kavukcuoglu, Rémi Munos, and Michal Valko.
\newblock Bootstrap your own latent: A new approach to self-supervised learning, 2020.

\bibitem[He et~al.(2021)He, Chen, Liu, Kortylewski, Yang, Bai, Wang, and Yuille]{transFG}
Ju~He, Jieneng Chen, Shuai Liu, Adam Kortylewski, Cheng Yang, Yutong Bai, Changhu Wang, and Alan~Loddon Yuille.
\newblock Transfg: A transformer architecture for fine-grained recognition.
\newblock In \emph{AAAI Conference on Artificial Intelligence}, 2021.
\newblock URL \url{https://api.semanticscholar.org/CorpusID:232233178}.

\bibitem[He et~al.(2016)He, Zhang, Ren, and Sun]{he2016deep}
Kaiming He, Xiangyu Zhang, Shaoqing Ren, and Jian Sun.
\newblock Deep residual learning for image recognition.
\newblock In \emph{Proceedings of the IEEE conference on computer vision and pattern recognition}, pages 770--778, 2016.

\bibitem[Hoffer et~al.(2019)Hoffer, Ben-Nun, Hubara, Giladi, Hoefler, and Soudry]{hoffer2019augment}
Elad Hoffer, Tal Ben-Nun, Itay Hubara, Niv Giladi, Torsten Hoefler, and Daniel Soudry.
\newblock Augment your batch: better training with larger batches.
\newblock \emph{arXiv preprint arXiv:1901.09335}, 2019.

\bibitem[Horn and Perona(2017)]{horn2017devil}
Grant~van Horn and Pietro Perona.
\newblock The devil is in the tails: Fine-grained classification in the wild.
\newblock \emph{arXiv preprint arXiv:1709.01450}, 2, 2017.

\bibitem[Huang et~al.(2016)Huang, Liu, and Weinberger]{Huang2016DenselyCC}
Gao Huang, Zhuang Liu, and Kilian~Q. Weinberger.
\newblock Densely connected convolutional networks.
\newblock \emph{2017 IEEE Conference on Computer Vision and Pattern Recognition (CVPR)}, pages 2261--2269, 2016.
\newblock URL \url{https://api.semanticscholar.org/CorpusID:9433631}.

\bibitem[Hui and Belkin(2021)]{hui2021evaluation}
Like Hui and Mikhail Belkin.
\newblock Evaluation of neural architectures trained with square loss vs cross-entropy in classification tasks, 2021.

\bibitem[Jaiswal et~al.(2021)Jaiswal, Babu, Zadeh, Banerjee, and Makedon]{jaiswal2021survey}
Ashish Jaiswal, Ashwin~Ramesh Babu, Mohammad~Zaki Zadeh, Debapriya Banerjee, and Fillia Makedon.
\newblock A survey on contrastive self-supervised learning, 2021.

\bibitem[Kim et~al.(2021)Kim, Oh, Kim, Cho, and Yun]{kim2021comparing}
Taehyeon Kim, Jaehoon Oh, NakYil Kim, Sangwook Cho, and Se-Young Yun.
\newblock Comparing kullback-leibler divergence and mean squared error loss in knowledge distillation, 2021.

\bibitem[Krause et~al.(2013)Krause, Deng, Stark, and Fei-Fei]{Krause2013CollectingAL}
Jonathan Krause, Jia Deng, Michael Stark, and Li~Fei-Fei.
\newblock Collecting a large-scale dataset of fine-grained cars.
\newblock 2013.
\newblock URL \url{https://api.semanticscholar.org/CorpusID:16632981}.

\bibitem[Kurtulus et~al.(2023)Kurtulus, Li, Dauphin, and Cubuk]{10.5555/3618408.3619150}
Emirhan Kurtulus, Zichao Li, Yann Dauphin, and Ekin~D. Cubuk.
\newblock Tied-augment: controlling representation similarity improves data augmentation.
\newblock In \emph{Proceedings of the 40th International Conference on Machine Learning}, ICML'23. JMLR.org, 2023.

\bibitem[Lagunas et~al.(2023)Lagunas, Impata, Martinez, Fernandez, Georgakis, Braun, and Bertrand]{lagunas2023transfer}
Manuel Lagunas, Brayan Impata, Victor Martinez, Virginia Fernandez, Christos Georgakis, Sofia Braun, and Felipe Bertrand.
\newblock Transfer learning for fine-grained classification using semi-supervised learning and visual transformers.
\newblock \emph{arXiv preprint arXiv:2305.10018}, 2023.

\bibitem[Lin et~al.(2015)Lin, RoyChowdhury, and Maji]{7410527}
Tsung-Yu Lin, Aruni RoyChowdhury, and Subhransu Maji.
\newblock Bilinear cnn models for fine-grained visual recognition.
\newblock In \emph{2015 IEEE International Conference on Computer Vision (ICCV)}, pages 1449--1457, 2015.
\newblock \doi{10.1109/ICCV.2015.170}.

\bibitem[Lu et~al.(2022)Lu, Lu, Yu, and Wang]{LU202225}
Ziqian Lu, Zheming Lu, Yunlong Yu, and Zonghui Wang.
\newblock Learn more from less: Generalized zero-shot learning with severely limited labeled data.
\newblock \emph{Neurocomputing}, 477:\penalty0 25--35, 2022.
\newblock ISSN 0925-2312.
\newblock \doi{https://doi.org/10.1016/j.neucom.2022.01.007}.
\newblock URL \url{https://www.sciencedirect.com/science/article/pii/S0925231222000078}.

\bibitem[Maji et~al.(2013)Maji, Kannala, Rahtu, Blaschko, and Vedaldi]{maji13fine-grained}
S.~Maji, J.~Kannala, E.~Rahtu, M.~Blaschko, and A.~Vedaldi.
\newblock Fine-grained visual classification of aircraft.
\newblock Technical report, 2013.

\bibitem[Mao et~al.(2023)Mao, Mohri, and Zhong]{mao2023crossentropy}
Anqi Mao, Mehryar Mohri, and Yutao Zhong.
\newblock Cross-entropy loss functions: Theoretical analysis and applications, 2023.

\bibitem[Mazumder et~al.(2021)Mazumder, Singh, and Namboodiri]{mazumder2021fair}
Pratik Mazumder, Pravendra Singh, and Vinay~P. Namboodiri.
\newblock Fair visual recognition in limited data regime using self-supervision and self-distillation, 2021.

\bibitem[Radford et~al.(2021)Radford, Kim, Hallacy, Ramesh, Goh, Agarwal, Sastry, Askell, Mishkin, Clark, et~al.]{radford2021learning}
Alec Radford, Jong~Wook Kim, Chris Hallacy, Aditya Ramesh, Gabriel Goh, Sandhini Agarwal, Girish Sastry, Amanda Askell, Pamela Mishkin, Jack Clark, et~al.
\newblock Learning transferable visual models from natural language supervision.
\newblock In \emph{International conference on machine learning}, pages 8748--8763. PMLR, 2021.

\bibitem[Ramdan et~al.(2020)Ramdan, Heryana, Arisal, Kusumo, and Pardede]{9298575}
Ade Ramdan, Ana Heryana, Andria Arisal, R.~Budiarianto~S. Kusumo, and Hilman~F. Pardede.
\newblock Transfer learning and fine-tuning for deep learning-based tea diseases detection on small datasets.
\newblock In \emph{2020 International Conference on Radar, Antenna, Microwave, Electronics, and Telecommunications (ICRAMET)}, pages 206--211, 2020.

\bibitem[Schmarje et~al.(2021)Schmarje, Santarossa, Schr{\"o}der, and Koch]{schmarje2021survey}
Lars Schmarje, Monty Santarossa, Simon-Martin Schr{\"o}der, and Reinhard Koch.
\newblock A survey on semi-, self-and unsupervised learning for image classification.
\newblock \emph{IEEE Access}, 9:\penalty0 82146--82168, 2021.

\bibitem[Shu et~al.(2022)Shu, Yu, Xu, and Liu]{shu2022improving}
Yangyang Shu, Baosheng Yu, Haiming Xu, and Lingqiao Liu.
\newblock Improving fine-grained visual recognition in low data regimes via self-boosting attention mechanism.
\newblock In \emph{European Conference on Computer Vision}, pages 449--465. Springer, 2022.

\bibitem[Simonyan and Zisserman(2014)]{simonyan2014very}
Karen Simonyan and Andrew Zisserman.
\newblock Very deep convolutional networks for large-scale image recognition.
\newblock \emph{arXiv preprint arXiv:1409.1556}, 2014.

\bibitem[Sultana et~al.(2022)Sultana, Naseer, Khan, Khan, and Khan]{sultana2022selfdistilled}
Maryam Sultana, Muzammal Naseer, Muhammad~Haris Khan, Salman Khan, and Fahad~Shahbaz Khan.
\newblock Self-distilled vision transformer for domain generalization, 2022.

\bibitem[Szegedy et~al.(2015)Szegedy, Liu, Jia, Sermanet, Reed, Anguelov, Erhan, Vanhoucke, and Rabinovich]{7298594}
C.~Szegedy, Wei Liu, Yangqing Jia, P.~Sermanet, S.~Reed, D.~Anguelov, D.~Erhan, V.~Vanhoucke, and A.~Rabinovich.
\newblock Going deeper with convolutions.
\newblock In \emph{2015 IEEE Conference on Computer Vision and Pattern Recognition (CVPR)}, pages 1--9, Los Alamitos, CA, USA, jun 2015. IEEE Computer Society.
\newblock \doi{10.1109/CVPR.2015.7298594}.
\newblock URL \url{https://doi.ieeecomputersociety.org/10.1109/CVPR.2015.7298594}.

\bibitem[Szegedy et~al.(2016)Szegedy, Vanhoucke, Ioffe, Shlens, and Wojna]{7780677}
Christian Szegedy, Vincent Vanhoucke, Sergey Ioffe, Jon Shlens, and Zbigniew Wojna.
\newblock Rethinking the inception architecture for computer vision.
\newblock In \emph{2016 IEEE Conference on Computer Vision and Pattern Recognition (CVPR)}, pages 2818--2826, 2016.
\newblock \doi{10.1109/CVPR.2016.308}.

\bibitem[Tang et~al.(2022)Tang, Yuan, Li, and Tang]{tang2022learning}
Hao Tang, Chengcheng Yuan, Zechao Li, and Jinhui Tang.
\newblock Learning attention-guided pyramidal features for few-shot fine-grained recognition.
\newblock \emph{Pattern Recognition}, 130:\penalty0 108792, 2022.

\bibitem[Touvron et~al.(2021)Touvron, Cord, Douze, Massa, Sablayrolles, and Jegou]{deit}
Hugo Touvron, Matthieu Cord, Matthijs Douze, Francisco Massa, Alexandre Sablayrolles, and Herve Jegou.
\newblock Training data-efficient image transformers \& distillation through attention.
\newblock In Marina Meila and Tong Zhang, editors, \emph{Proceedings of the 38th International Conference on Machine Learning}, volume 139 of \emph{Proceedings of Machine Learning Research}, pages 10347--10357. PMLR, 18--24 Jul 2021.
\newblock URL \url{https://proceedings.mlr.press/v139/touvron21a.html}.

\bibitem[Touvron et~al.(2022)Touvron, Cord, and Jégou]{DeiT-III-Revenge-of-the-ViT}
Hugo Touvron, Matthieu Cord, and Hervé Jégou.
\newblock Deit iii: Revenge of the vit, 2022.

\bibitem[Wah et~al.(2011)Wah, Branson, Welinder, Perona, and Belongie]{WahCUB_200_2011}
C.~Wah, S.~Branson, P.~Welinder, P.~Perona, and S.~Belongie.
\newblock Technical report. the caltech-ucsd birds-200-2011 dataset.
\newblock Technical Report CNS-TR-2011-001, California Institute of Technology, 2011.

\bibitem[Wang et~al.(2021)Wang, Yu, and Gao]{Wang2021FeatureFV}
Jun Wang, Xiaohan Yu, and Yongsheng Gao.
\newblock Feature fusion vision transformer for fine-grained visual categorization.
\newblock In \emph{British Machine Vision Conference}, 2021.
\newblock URL \url{https://api.semanticscholar.org/CorpusID:235742913}.

\bibitem[Wang and Qi(2022)]{wang2022contrastive}
Xiao Wang and Guo-Jun Qi.
\newblock Contrastive learning with stronger augmentations.
\newblock \emph{IEEE transactions on pattern analysis and machine intelligence}, 45\penalty0 (5):\penalty0 5549--5560, 2022.

\bibitem[Wang et~al.(2022)Wang, Fan, Tian, Kihara, and Chen]{wang2022importance}
Xiao Wang, Haoqi Fan, Yuandong Tian, Daisuke Kihara, and Xinlei Chen.
\newblock On the importance of asymmetry for siamese representation learning.
\newblock In \emph{Proceedings of the IEEE/CVF Conference on Computer Vision and Pattern Recognition}, pages 16570--16579, 2022.

\bibitem[Xu et~al.(2021)Xu, Zhang, Hu, Wang, Wang, Wei, Bai, and Liu]{xu2021end}
Mengde Xu, Zheng Zhang, Han Hu, Jianfeng Wang, Lijuan Wang, Fangyun Wei, Xiang Bai, and Zicheng Liu.
\newblock End-to-end semi-supervised object detection with soft teacher.
\newblock In \emph{Proceedings of the IEEE/CVF International Conference on Computer Vision}, pages 3060--3069, 2021.

\bibitem[Yang et~al.(2022)Yang, Song, King, and Xu]{yang2022survey}
Xiangli Yang, Zixing Song, Irwin King, and Zenglin Xu.
\newblock A survey on deep semi-supervised learning.
\newblock \emph{IEEE Transactions on Knowledge and Data Engineering}, 2022.

\bibitem[Yu et~al.(2018)Yu, Zhao, Zheng, Zhang, and You]{10.1007/978-3-030-01270-0_35}
Chaojian Yu, Xinyi Zhao, Qi~Zheng, Peng Zhang, and Xinge You.
\newblock Hierarchical bilinear pooling for fine-grained visual recognition.
\newblock In Vittorio Ferrari, Martial Hebert, Cristian Sminchisescu, and Yair Weiss, editors, \emph{Computer Vision -- ECCV 2018}, pages 595--610, Cham, 2018. Springer International Publishing.
\newblock ISBN 978-3-030-01270-0.

\bibitem[Yun et~al.(2019)Yun, Han, Oh, Chun, Choe, and Yoo]{yun2019cutmix}
Sangdoo Yun, Dongyoon Han, Seong~Joon Oh, Sanghyuk Chun, Junsuk Choe, and Youngjoon Yoo.
\newblock Cutmix: Regularization strategy to train strong classifiers with localizable features.
\newblock In \emph{Proceedings of the IEEE/CVF international conference on computer vision}, pages 6023--6032, 2019.

\bibitem[Zbontar et~al.(2021)Zbontar, Jing, Misra, LeCun, and Deny]{zbontar2021barlow}
Jure Zbontar, Li~Jing, Ishan Misra, Yann LeCun, and Stéphane Deny.
\newblock Barlow twins: Self-supervised learning via redundancy reduction, 2021.

\bibitem[Zhang et~al.(2014)Zhang, Donahue, Girshick, and Darrell]{zhang2014part}
Ning Zhang, Jeff Donahue, Ross Girshick, and Trevor Darrell.
\newblock Part-based r-cnns for fine-grained category detection.
\newblock In \emph{Computer Vision--ECCV 2014: 13th European Conference, Zurich, Switzerland, September 6-12, 2014, Proceedings, Part I 13}, pages 834--849. Springer, 2014.

\bibitem[Zhang and Sabuncu(2018)]{zhang2018generalized}
Zhilu Zhang and Mert~R. Sabuncu.
\newblock Generalized cross entropy loss for training deep neural networks with noisy labels, 2018.

\bibitem[Zheng et~al.(2019)Zheng, Fu, Zha, and Luo]{10.5555/3454287.3454672}
Heliang Zheng, Jianlong Fu, Zheng-Jun Zha, and Jiebo Luo.
\newblock \emph{Learning deep bilinear transformation for fine-grained image representation}.
\newblock Curran Associates Inc., Red Hook, NY, USA, 2019.

\bibitem[Zhuang et~al.(2020)Zhuang, Wang, and Qiao]{zhuang2020learning}
Peiqin Zhuang, Yali Wang, and Yu~Qiao.
\newblock Learning attentive pairwise interaction for fine-grained classification.
\newblock In \emph{Proceedings of the AAAI conference on artificial intelligence}, volume~34, pages 13130--13137, 2020.

\end{thebibliography}

\clearpage
\appendix

\setcounter{page}{1}

\section{Additional Field Overview}
\label{app:detailed_related}

\subsection{Efficient Learning} 

\textbf{Augmentations.}
Augmentation techniques are crucial in advancing deep learning models \cite{wang2022importance}, providing strategies that enhance training efficiency, improve generalisation, and bolster model robustness. 
One of the recent groundworks includes batch augmentation \cite{hoffer2019augment}, which is a powerful strategy that utilises large batches comprising multiple transformations for each sample. This not only accelerates training by reducing the number of stochastic gradient descent (SGD) updates but also acts as a regulariser, leading to improved generalisation.
Another popular approach is RandAugment \cite{cubuk2020randaugment}, designed for automated data augmentation by narrowing the search space and providing parameterisation. This method consistently outperforms previous automated augmentation techniques, demonstrating its efficiency across various tasks and datasets.
Recent advances in contrastive learning also demonstrate that the so-called Stronger Augmentations \cite{wang2022contrastive} significantly enhance contrastive learning by enforcing distributional divergence between images augmented with such "strong" permutations as random cropping and flipping.
Addressing limitations associated with regional dropout strategies, CutMix  \cite{yun2019cutmix} involves cutting and pasting patches among training images. This ensures information preservation, promotes object localisation capabilities
and improves the model's resilience against input corruption.

\section{Experimental Details}

\subsection{Datasets}
\label{app:datasets}

\begin{table}[!ht]
  \caption{The details of three fine-grained visual classification datasets used for the experiments.}
  \label{tab:datasets}
  \vskip 0.03in
  \centering
  \begin{tabular}{l|c|c|c}
    \toprule
    Dataset & Categories & Classes & Images  \\
    \midrule
    CUB-200-2011 \cite{WahCUB_200_2011} 
    & Birds & 200 & 11,788 \\
    Stanford Cars \cite{Krause2013CollectingAL} 
    & Cars & 196 & 16,185 \\
    FGVC-Aircraft \cite{maji13fine-grained} 
    & Airplanes & 102  & 10,200 \\
  \bottomrule
\end{tabular}
\end{table}

\subsection{Implementation Details}
\label{app:implementation}

For our self-distillation part, after performing random sampling from the input image, we resize both target and source sampled regions to $224\times224$ pixels. The motivation is to provide more different scales for input images so that the model can learn more scale-invariant representations, which are usually assumed to be already present in standard-sized datasets.
Our training setup includes the standard SGD optimiser with a momentum equal to 0.9, a learning rate of 0.03, and a training batch size of 24 for all datasets.
All experiments have been conducted on a single NVIDIA RTX 6000 GPU using the PyTorch framework and the APEX utility for mixed precision training.

\section{Additional Analysis}

\subsection{Qualitative Analysis}
\label{app:anal_qual}

\begin{figure}[!h]
    \centering
    \vskip -0.06in
    \includegraphics[width=0.49\linewidth]{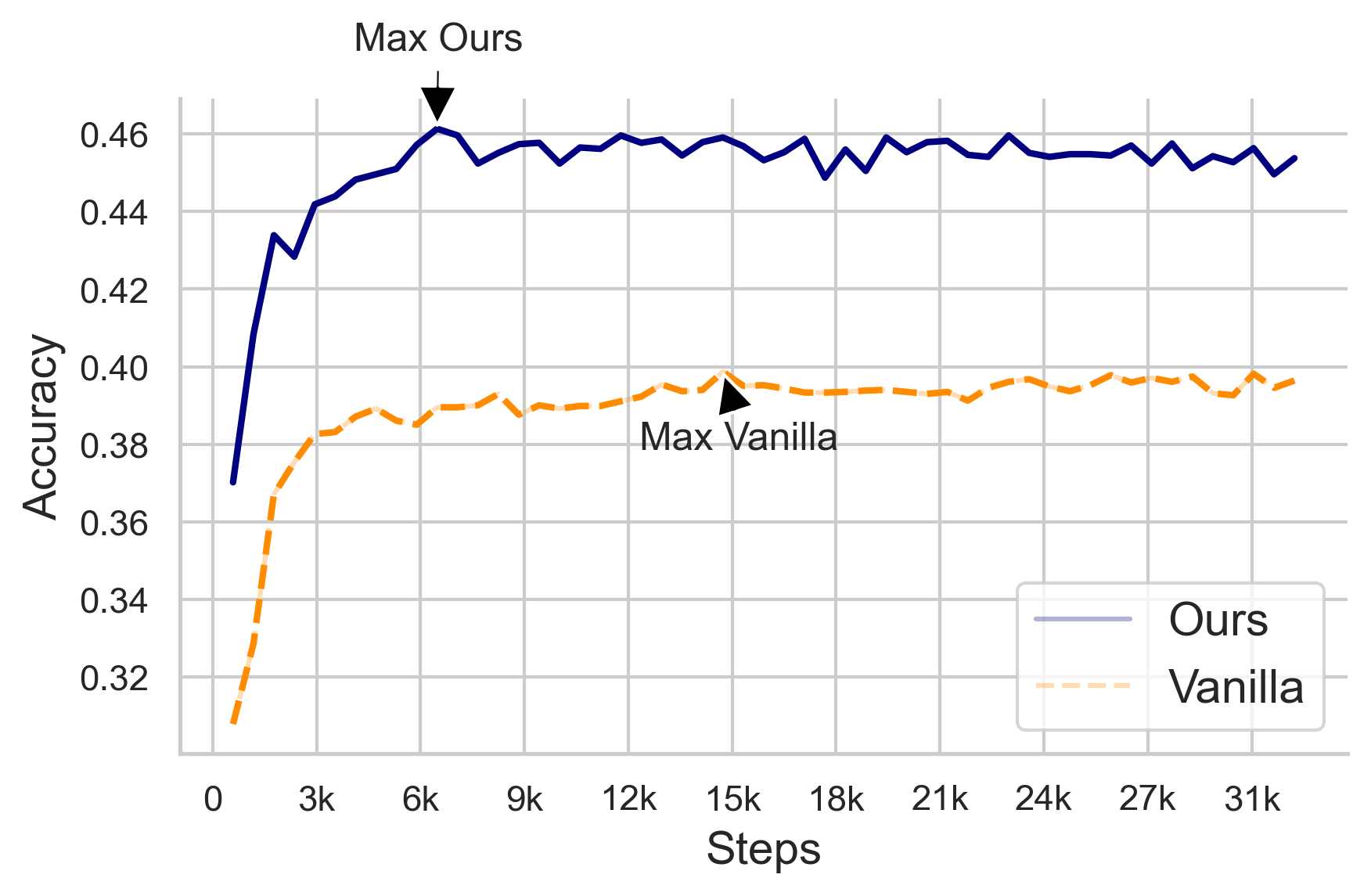}
    \includegraphics[width=0.49\linewidth]{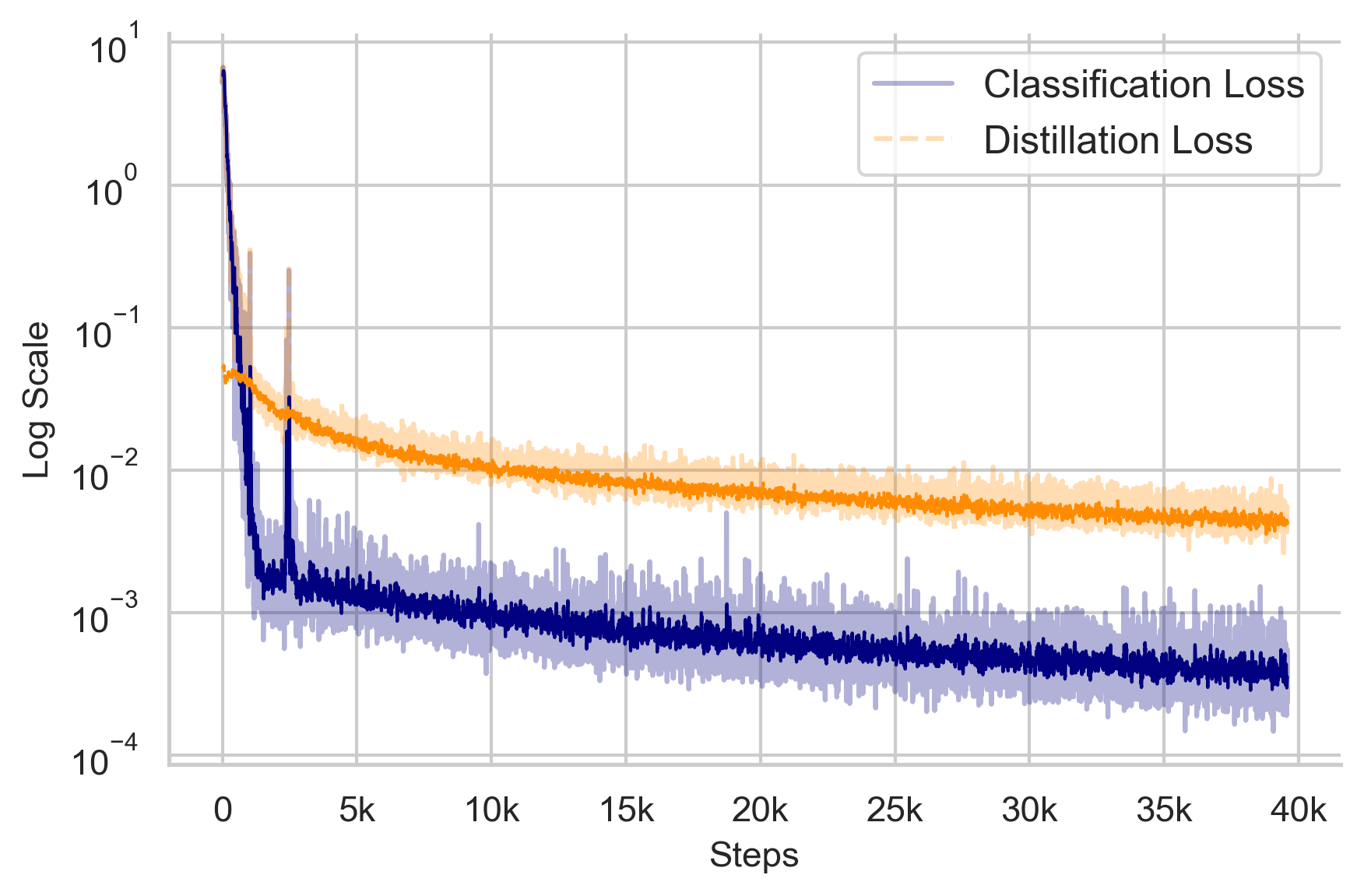}

    \vskip -0.12in
    \caption{\textit{Left}: Comparison of test accuracy evolution for our approach
    and traditionally fine-tuned vanilla ResNet-50 
   (CUB 10 \% dataset).
   \textit{Right}: Training loss evolution for both classification and distillation objectives in our AD-Net (ResNet-50, CUB 10 \% dataset).}
    \label{fig:accuracy_loss}
\end{figure}

In order to analyse the motivation behind the significant performance improvement with our approach, we provide a direct comparison of training behaviour for a fine-tuned vanilla ResNet-50 and our self-distilled AD-Net with the same backbone.
It can be observed from Figure \ref{fig:accuracy_loss} (left) that the model trained with standard fine-tuning procedure tends to overfit the limited data samples quickly and does not allow further refinement, while our framework is able to avoid overfitting more effectively.
This can be explained by the effect of the additional distillation objective, which introduces an independent and more detailed source of information by enforcing feature space alignment for augmented views of the same image.

This specific effect is demonstrated in Figure \ref{fig:accuracy_loss} (right), where we provide the evolution of both classification and distillation objectives separately. Specifically, categorisation loss gets saturated quickly and becomes insignificant after the first 10\% of training time, while our self-distillation component provides a noticeable effect throughout the whole training process. 
This is especially useful in low-data regimes, where models tend to quickly overfit to the main classification loss and do not obtain significant update signals, which is not the case for our proposed multi-component objective function.

\subsection{Ablation Study}
\label{app:anal_ablation}

\noindent\textbf{Distillation Loss.}
To rigorously evaluate the impact of distillation loss within our framework, a series of experiments were conducted on the CUB 10\% dataset. 
First, in order to find the most effective type of distillation loss, we explore various objective functions applied at different type of outputs. 
In Table \ref{tab:losses} we illustrate the variance in performance among different loss functions when 
applied to measure the disparity between features or logits 
coming from the target and source distillation branches in our architecture. 
Notably, Cross Entropy and Focal Loss functions, when applied to the output logits of both branches, demonstrate inferior performance. 
While computing the loss on the output features (converted to normalised distributions with softmax) 
is more effective, with KL divergence showing best results compared to the L1 and L2 objectives. One plausible explanation for the superior efficacy of KL divergence over other loss functions could be its ability to compare more sophisticated abstractions coming from
differently augmented views.

\begin{table}[!ht]
    \centering
      \caption{Effects caused by the type of distillation loss on the final metric. KL divergence computed between feature outputs of distillation branches shows the best 
      performance.}
      \label{tab:losses}
      \vskip 0.03in
    \begin{tabular}{l@{\hspace{2em}}|c|c}

    \toprule
    Loss & Features & Logits \\
    \midrule

    Cross Entropy & - & 39.09 \\
    Focal Loss & - & 39.32 \\ 
    L1 (MAE) & 43.96 & - \\
    L2 (MSE) & 41.67 & - \\     
    Kullback–Leibler (KL) & \textbf{47.51}
    & - \\
    \bottomrule
  \end{tabular}
    \vskip -0.02in
\end{table}

Following the identification of the most effective loss function for our distillation branch, we proceeded to investigate its influence on the aggregated loss, denoted as $\mathcal{L}_{agg}$ in Eq. \ref{eq:final_loss}, by varying the weight coefficient $\alpha$. 
In order to find the most optimal value, we experiment
with both constant and variable $\alpha$ values, and
summarise the results in Table \ref{tab:alph}. Our heuristic findings revealed that the model achieves its peak performance with the value of $\alpha$ set to 0.1.

\begin{table}[!h]
    \centering
    \caption{Study of the effect of $\alpha$ coefficient in the aggregated loss $\mathcal{L}_{agg}$ from Eq. \ref{eq:final_loss}. The $\alpha$ decay is a linear decay from 1.0 to 0.01 .}
    \label{tab:alph} 
    \vskip 0.03in
    \begin{tabular}{l|c|c|c|c|c}
    \toprule
    $\alpha$   & 1 & 0.5 & 0.1 & 0.01 & $\alpha$ decay \\
    \midrule
    Acc, \%  &  45.98  &  46.13 &  \textbf{47.51}
    &  44.08  &  45.6 \\
    \bottomrule
    \end{tabular}
    \vskip -0.1in
\end{table}

\subsection{Limitations}
\label{app:limitations}

Although our approach demonstrates a significant performance gain in the low-data setting, we also analyse and acknowledge its current limitations.
First, due to the extra forward passes for each of the separate branches, the training time is increased by approximately 35-90 \% compared to the vanilla fine-tuning (depending on the architecture type, see \mbox{Tab. \ref{tab:ablation_architecture}}).
However, this matter is not presented at inference time since our solution brings zero compute and time cost after training.
Second, our approach has an inversely proportional performance gain to the size and diversity of a dataset (refer to Table \ref{tab:results_main}), which theoretically may lead to less significant accuracy improvement when the data is abundant.
Lastly, our solution requires a hyper-parameter $\alpha$ in Eq. \ref{eq:final_loss} for controlling the influence of the distillation objective function on the overall loss, which unadapted value may sometimes lead to unstable training results 
(due to the nature of the KL divergence loss).
We suggest that our current heuristic choice can be potentially replaced by an independent learnable parameter.

\subsection{Quantitative Analysis}

\subsubsection{State-of-the-art comparison}
\label{app:results_full}

\textbf{Main Results.} In Table \ref{tab:results_full} we provide the full comparison of different approaches on all data percentages including full datasets. 

\noindent\textbf{Other Results.} Additionally, in Table \ref{tab:other_models}, we also compare other methods potentially suitable for the low-data setting. 
Namely, SwAV \cite{swav}, pre-trained in a self-supervised way and fine-tuned, and CLIP \cite{radford2021learning}, a self-supervised vision-language model, used for zero-shot inference. As can be seen, compared to our AD-Net, they demonstrate promising but unstable results.

\begin{table*}[!ht]
\centering
\caption{Comparison of different approaches 
using various percentages of the data on three popular FGIC datasets. Our proposed solution achieves consistent improvement in performance over other methods across low data settings. 
Best results are highlighted in bold.}
\label{tab:results_full}
\vskip 0.03in
\begin{tabular}{@{}c|l|ccccc@{}}
\toprule
\multirow{2}{*}{Dataset} &
  \multicolumn{1}{c|}{\multirow{2}{*}{Method}} &
  \multicolumn{5}{c}{Training data percentage} \\ \cmidrule(l){3-7} 
 &
  \multicolumn{1}{c|}{} &
  \multicolumn{1}{@{\hspace{1em}}c@{\hspace{1em}}|}{10\%} &
  \multicolumn{1}{@{\hspace{1em}}c@{\hspace{1em}}|}{15\%} &
  \multicolumn{1}{@{\hspace{1em}}c@{\hspace{1em}}|}{30\%} &
  \multicolumn{1}{@{\hspace{1em}}c@{\hspace{1em}}|}{50\%} &
  {\hspace{0.5em}}{{\color[HTML]{9B9B9B} 100\%}}{\hspace{0.5em}} \\ \midrule
\multirow{3}{*}{CUB-200-2011 
} &
  ResNet-50
  &
  \multicolumn{1}{c|}{36.99} &
  \multicolumn{1}{c|}{48.88} &
  \multicolumn{1}{c|}{62.60} &
  \multicolumn{1}{c|}{73.23} &
  {\color[HTML]{9B9B9B} 81.34} \\
 &
  FBP
  &
  \multicolumn{1}{c|}{37.88} &
  \multicolumn{1}{c|}{49.12} &
  \multicolumn{1}{c|}{63.27} &
  \multicolumn{1}{c|}{73.70} &
  {\color[HTML]{9B9B9B} 82.52} \\
 &
  CBP-TS
  &
  \multicolumn{1}{c|}{37.12} &
  \multicolumn{1}{c|}{47.82} &
  \multicolumn{1}{c|}{62.24} &
  \multicolumn{1}{c|}{72.37} &
  {\color[HTML]{9B9B9B} 81.48} \\
 &
  HBP
  &
  \multicolumn{1}{c|}{38.57} &
  \multicolumn{1}{c|}{50.12} &
  \multicolumn{1}{c|}{63.86} &
  \multicolumn{1}{c|}{74.18} &
  {\color[HTML]{9B9B9B} \textbf{86.12}} \\
 &
  DBTNet-50 
  &
  \multicolumn{1}{c|}{37.67} &
  \multicolumn{1}{c|}{49.52} &
  \multicolumn{1}{c|}{63.16} &
  \multicolumn{1}{c|}{73.28} &
  {\color[HTML]{9B9B9B} 86.04} \\  
 &
  SAM (ResNet-50)
  &
  \multicolumn{1}{c|}{40.24} &
  \multicolumn{1}{c|}{52.05} &
  \multicolumn{1}{c|}{64.07} &
  \multicolumn{1}{c|}{73.92} &
  {\color[HTML]{9B9B9B} 81.62} \\
 &
  SAM (FBP)
  &
  \multicolumn{1}{c|}{41.83} &
  \multicolumn{1}{c|}{52.35} &
  \multicolumn{1}{c|}{65.19} &
  \multicolumn{1}{c|}{74.54} &
  {\color[HTML]{9B9B9B} 81.86} \\  
 &
  Ours (ResNet-50) &
  \multicolumn{1}{c|}{\textbf{47.51}} &
  \multicolumn{1}{c|}{\textbf{60.08}} &
  \multicolumn{1}{c|}{\textbf{71.11}} &
  \multicolumn{1}{c|}{\textbf{77.67}} &
  {\color[HTML]{9B9B9B} 82.06} \\ \midrule 
\multirow{3}{*}{Stanford Cars
} &
  ResNet-50
  &
  \multicolumn{1}{c|}{37.45} &
  \multicolumn{1}{c|}{53.01} &
  \multicolumn{1}{c|}{75.26} &
  \multicolumn{1}{c|}{83.56} &
  {\color[HTML]{9B9B9B} 91.02} \\
 &
  FBP
  &
  \multicolumn{1}{c|}{40.13} &
  \multicolumn{1}{c|}{55.07} &
  \multicolumn{1}{c|}{76.42} &
  \multicolumn{1}{c|}{85.10} &
  {\color[HTML]{9B9B9B} 91.63} \\
 &
  CBP-TS
  &
  \multicolumn{1}{c|}{37.77} &
  \multicolumn{1}{c|}{54.87} &
  \multicolumn{1}{c|}{75.51} &
  \multicolumn{1}{c|}{84.80} &
  {\color[HTML]{9B9B9B} 89.52} \\
 &
  HBP
  &
  \multicolumn{1}{c|}{40.02} &
  \multicolumn{1}{c|}{55.82} &
  \multicolumn{1}{c|}{76.81} &
  \multicolumn{1}{c|}{85.31} &
  {\color[HTML]{9B9B9B} 92.73} \\
 &
  DBTNet-50 
  &
  \multicolumn{1}{c|}{39.48} &
  \multicolumn{1}{c|}{55.24} &
  \multicolumn{1}{c|}{76.52} &
  \multicolumn{1}{c|}{86.52} &
  {\color[HTML]{9B9B9B} \textbf{94.32}} \\  
 &
  SAM (ResNet-50)
  &
  \multicolumn{1}{c|}{39.96} &
  \multicolumn{1}{c|}{55.02} &
  \multicolumn{1}{c|}{76.69} &
  \multicolumn{1}{c|}{84.85} &
  {\color[HTML]{9B9B9B} 91.06} \\
 &
  SAM (FBP)  
  &
  \multicolumn{1}{c|}{43.19} &
  \multicolumn{1}{c|}{57.42} &
  \multicolumn{1}{c|}{77.63} &
  \multicolumn{1}{c|}{85.71} &
  {\color[HTML]{9B9B9B} 91.48} \\  
 &
  Ours (ResNet-50) &
  \multicolumn{1}{c|}{\textbf{55.09}} &
  \multicolumn{1}{c|}{\textbf{67.42}} &
  \multicolumn{1}{c|}{\textbf{81.53}} &
  \multicolumn{1}{c|}{\textbf{87.41}} &
  {\color[HTML]{9B9B9B} 91.96} \\ \midrule 
\multirow{3}{*}{FGVC-Aircraft
} &
  ResNet-50
  &
  \multicolumn{1}{c|}{43.52} &
  \multicolumn{1}{c|}{53.17} &
  \multicolumn{1}{c|}{71.32} &
  \multicolumn{1}{c|}{78.61} &
  {\color[HTML]{9B9B9B} 87.13} \\
 &
  FBP
  &
  \multicolumn{1}{c|}{45.16} &
  \multicolumn{1}{c|}{55.06} &
  \multicolumn{1}{c|}{72.12} &
  \multicolumn{1}{c|}{79.93} &
  {\color[HTML]{9B9B9B} 87.32} \\
 &
  CBP-TS
  &
  \multicolumn{1}{c|}{44.63} &
  \multicolumn{1}{c|}{54.79} &
  \multicolumn{1}{c|}{71.32} &
  \multicolumn{1}{c|}{79.60} &
  {\color[HTML]{9B9B9B} 84.58} \\
 &
  HBP
  &
  \multicolumn{1}{c|}{45.28} &
  \multicolumn{1}{c|}{56.12} &
  \multicolumn{1}{c|}{72.58} &
  \multicolumn{1}{c|}{81.47} &
  {\color[HTML]{9B9B9B} 89.74} \\
 &
  DBTNet-50 
  &
  \multicolumn{1}{c|}{45.35} &
  \multicolumn{1}{c|}{56.36} &
  \multicolumn{1}{c|}{73.06} &
  \multicolumn{1}{c|}{81.26} &
  {\color[HTML]{9B9B9B} \textbf{90.86}} \\  
 &
  SAM (ResNet-50) 
  &
  \multicolumn{1}{c|}{46.73} &
  \multicolumn{1}{c|}{56.02} &
  \multicolumn{1}{c|}{72.59} &
  \multicolumn{1}{c|}{79.21} &
  {\color[HTML]{9B9B9B} 86.74} \\
 &
  SAM (FBP) 
  &
  \multicolumn{1}{c|}{47.97} &
  \multicolumn{1}{c|}{57.47} &
  \multicolumn{1}{c|}{73.43} &
  \multicolumn{1}{c|}{80.86} &
  {\color[HTML]{9B9B9B} 87.46} \\  
 &
  Ours (ResNet-50) &
  \multicolumn{1}{c|}{\textbf{55.81}} &
  \multicolumn{1}{c|}{\textbf{62.59}} &
  \multicolumn{1}{c|}{\textbf{74.44}} &
  \multicolumn{1}{c|}{\textbf{81.73}} &
  {\color[HTML]{9B9B9B} 88.64} \\ \bottomrule 
\end{tabular}
\end{table*}

\begin{table}[!h]
\centering
\caption{Experiments with other models
on datasets with 10\% of training data.
Results for CLIP are obtained using zero-shot classification.
Vanilla and Our results are for comparison.
}
\label{tab:other_models}
\begin{tabular}{@{}l|l|@{\hspace{1em}}c@{\hspace{1em}}|@{\hspace{1em}}c@{\hspace{1em}}|@{\hspace{1em}}c@{\hspace{1em}}}
\toprule
Type                 & Method       & CUB   & Cars  & Air   \\ \midrule
\multirow{4}{*}{CNN}
                     & ResNet-50 (vanilla) \cite{he2016deep} 
                                          & 36.99 & 37.45 & 43.52 \\
                     & SwAV (ResNet-50) \cite{swav}  
                                          & 16.91 & 36.19 & 49.49 \\
                     & CLIP (ResNet-50, zero-shot) \cite{radford2021learning}  
                                          &   -   & 55.80 & 19.30 \\                                           
                     & Ours (ResNet-50)   & 47.51 & 55.09 & 55.81 \\ \midrule
\multirow{4}{*}{ViT} & ViT-B/32 (vanilla) \cite{dosovitskiy2020image} 
                                          & 65.60 & 28.21 & 33.84 \\
                     & TransFG (ViT-B/32) \cite{transFG}       
                                          & 64.91 &   -   &   -   \\
                     & CLIP (ViT-B/32, zero-shot) \cite{radford2021learning}  
                                          &   -   & 59.40 & 21.20 \\ 
                     & Ours (ViT-B/32)    & 69.27 & 33.34 & 36.01 \\ \bottomrule                     
\end{tabular}
\end{table}

\subsubsection{Transferability}
\label{app:transferability}

Additionally, in Table \ref{tab:transferability} we demonstrate the high transferability of our approach by utilising it on top of most of the popular CNN- and ViT-based backbones. The absolute improvement varies between 3-10 \% showcasing that distilling local augmentations of input images indeed promotes feature refinement and is practically an architecture-independed technique.

\begin{table}[!h]
\centering
\caption{Transferability study of AD-Net 
on the CUB dataset with 10\% of training data. 
Column $\Delta$ shows the absolute performance increase with our approach compared to a vanilla backbone.
}
\label{tab:transferability}
\begin{tabular}{@{}l|l|c|c|r@{}}
\toprule
Type                 & Backbone     & Vanilla & Ours  & $\Delta$      \\ \midrule
\multirow{7}{*}{CNN} & ResNet-18    & 34.79   & 41.39 & +6.60  \\
                     & ResNet-34    & 36.83   & 45.81 & +8.98  \\
                     & ResNet-50    & 36.99   & 47.51 & +10.52 \\
                     & ResNet-101    & 40.19   & 49.44 & +9.25 \\
                     & GoogleNet    & 33.32   & 39.11 & +5.79  \\
                     & Inception v3 & 40.16   & 44.88 & +4.72  \\
                     & DenseNet-169 & 41.42   & 49.13 & +7.71  \\ \midrule
\multirow{2}{*}{ViT} & ViT-B/32     & 65.60   & 69.27 & +3.67  \\
                     & FFVT B/32    & 65.79   & 68.13 & +2.34  \\ \bottomrule
\end{tabular}
\end{table}

\subsection{Augmentations with Naive Fine-tuning}
\label{app:extra_experiments}

Additionally, we conduct experiments with some advanced augmentation techniques, such as as ScaleMix \cite{wang2022importance}, MultiCrop \cite{caron2020unsupervised}, and AsymAug \cite{wang2022importance} applied in a naive way with standard fine-tuning. As can be observed in Table \ref{tab:augs_other}, without our distillation technique, the ResNet-50 performance with the advanced augmentations is below the baseline with basic augmentations. Where by the basic augmentations we assume the classical list of augmentations recommended in the literature for each dataset. This includes random cropping, colour jittering, random horizontal flip, and further normalisation.

\begin{table}[!h]
\centering
\caption{The results with existing augmentation techniques, such as ScaleMix, MultiCrop, and AsymAug applied in a naive way with standard fine-tuning. Advanced augmentations include more geometrical and colour-related perturbations from the popular AutoAugs approach. As can be observed, without our distillation technique, the performance is below the baseline ResNet-50 with basic augmentations. The results were obtained with 10\% low-data regime on the CUB training set.}

\vskip 0.03in
\begin{tabular}{@{}l|c|c@{}}
\toprule
\multicolumn{1}{c|}{\multirow{2}{*}{Augmentation}} &
  \multirow{2}{*}{\begin{tabular}[c]{@{}c@{}}Standard\\ procedure\end{tabular}} &
  \multirow{2}{*}{\begin{tabular}[c]{@{}c@{}}Our \\ procedure\end{tabular}} \\
\multicolumn{1}{c|}{} &       &                            \\ \midrule
Basic augmentations       & 36.99 & \multicolumn{1}{c}{40.05} \\
Advanced augmentations       & 27.82 & \multicolumn{1}{c}{30.68} \\
ScaleMix \cite{wang2022importance}              & 28.54 & \multicolumn{1}{c}{26.99} \\
MultiCrop \cite{caron2020unsupervised}            & 28.06 & \multicolumn{1}{c}{28.90} \\
AsymAugs \cite{wang2022importance}             & 30.29 & \multicolumn{1}{c}{30.96} \\
\bottomrule
\end{tabular}
\label{tab:augs_other}
\end{table}

We assume that more advanced and complex types of augmentations harm the learning process under low-data regimes, since the model may be unable to catch the locally important patterns due to harsh image perturbations.

\subsection{Activation Maps Visualisation}
\label{app:visuals}

\begin{figure}[!h]
    \centering
    \includegraphics[width=0.9\linewidth]{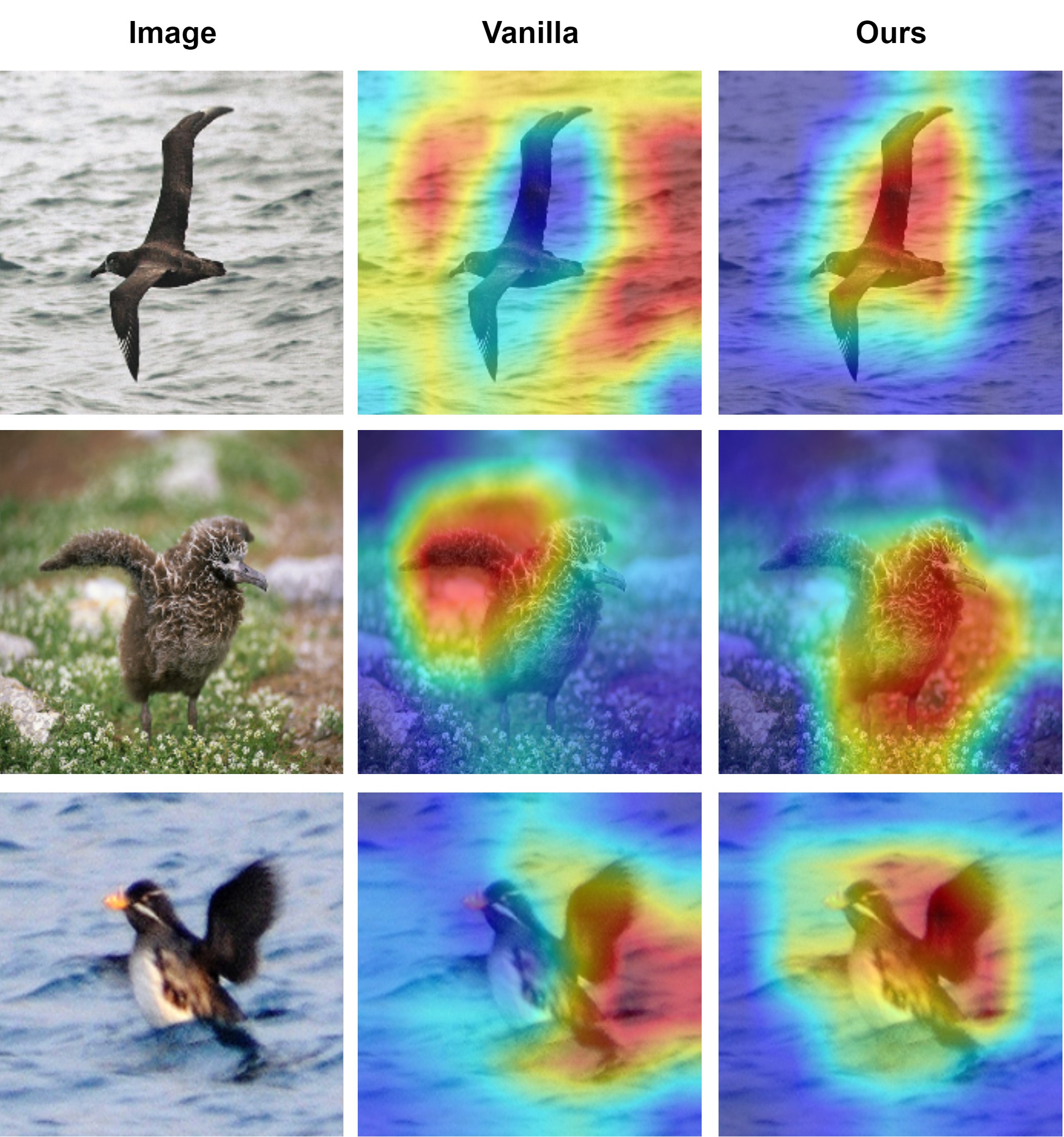}
    \caption{The visualisation of difference in feature activation maps between the vanilla ResNet-50 and our AD-Net on the CUB dataset. Red colour - higher activation, blue - lower activation.}
    \label{fig:vis_brids}
\end{figure}

\begin{figure}[!h]
    \centering
    \includegraphics[width=0.9\linewidth]{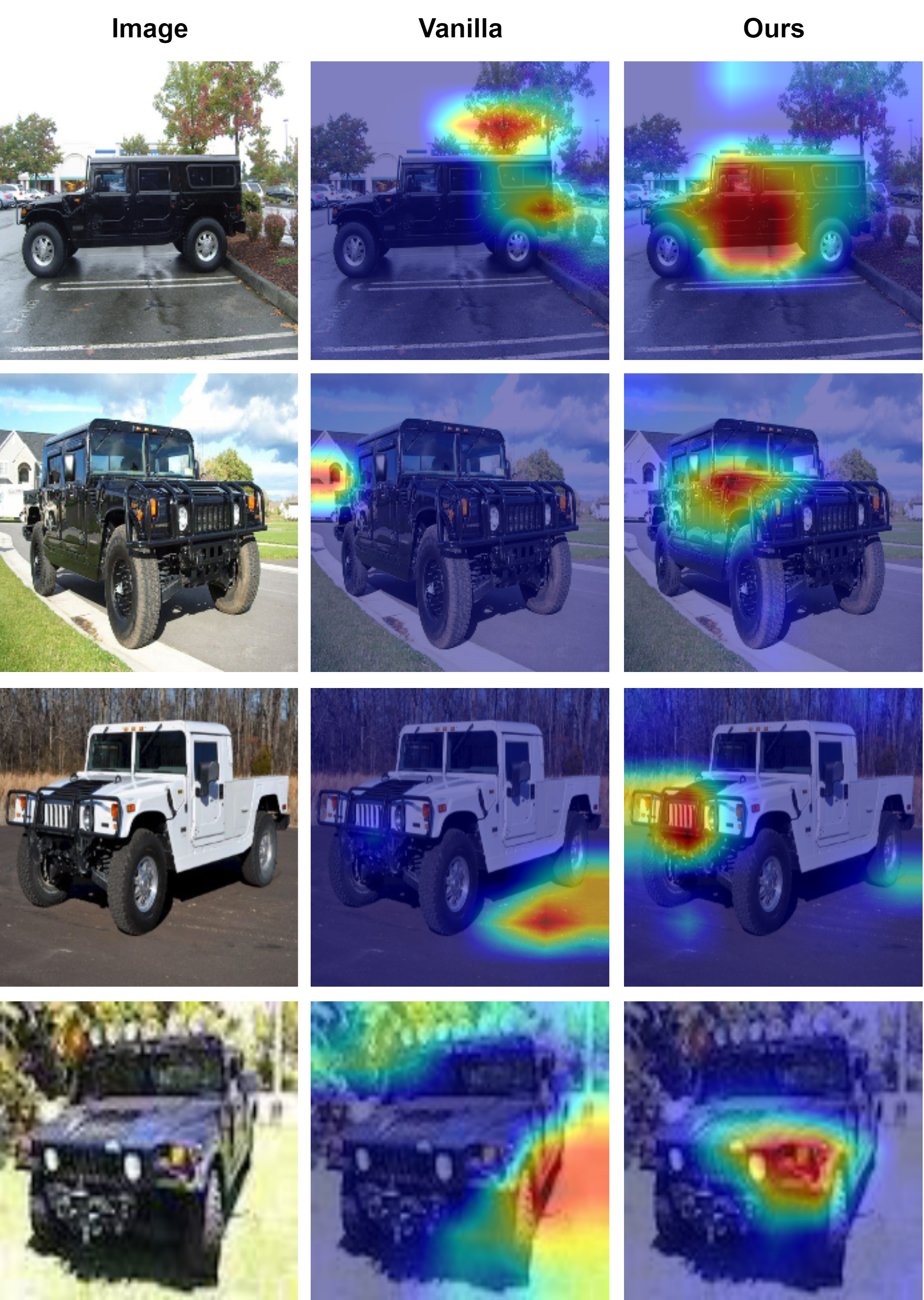}
    \caption{The visualisation of difference in feature activation maps between the vanilla ResNet-50 and our AD-Net on the Stanford Cars dataset. Red colour - higher activation, blue - lower activation.}
    \label{fig:vis_cars}
\end{figure}

In order to investigate the reason behind the significant performance improvement with our approach on the lowest data settings, we provide the difference in feature activation maps between the vanilla ResNet-50 and our AD-Net based on the same backbone.
In Figures \ref{fig:vis_brids} and \ref{fig:vis_cars} we can clearly observe higher quality of the activation area from our method (more attention to the distinctive foreground regions), which explains its noticeable performance gain.

\section{Application Guidelines}

\subsection{Answers for Potential Questions}
\label{app:qna}

\textbf{1. "How to decide whether should AD-Net be used for a given scenario?}

\textbf{A: }Our method is supposed to be used in the scenarious where the standard fine-tuning procedure shows poor performance due to a small amount of available labelled images.

\noindent\textbf{2. “How many images should be collected at least?”}

\textbf{A: } After a thorough investigation, we have concluded that the exact answer depends on multiple factors, such as the number of classes, the number of images per class, and the size and capacity of a chosen baseline.

Therefore, we believe some small prior experiments are needed to make the final decision. Specifically, we recommend starting with at least 5 images per class, and further increasing the amount until the minimum desired performance is achieved.

Although the general rule is “the more data - the better”, our solution was designed specifically for the cases where obtaining a lot of labelled data may be impractical.

\noindent\textbf{3. “When should AD-Net be switched to a different method as more training data are obtained?”}

\textbf{A: }We suggest tracking the overall performance increase compared to an initially chosen baseline along with the data increase. Once the performance gain reaches neglectable values a different method can be used instead.

However, our solution is targeting the cases where the total amount of labelled data is highly limited and obtaining more samples may be too difficult.

\end{document}